\documentclass[a4paper,fleqn]{cas-sc}  
\usepackage[numbers]{natbib}
\usepackage{amsfonts} 
\usepackage{algorithm}          
\usepackage{algpseudocode}  

\usepackage{array}                 
\usepackage{graphicx}     
\usepackage{multirow}   
\usepackage{url}   
\usepackage{hyperref} 
\usepackage{enumerate}  
\usepackage{float} 
\usepackage{balance} 
\hypersetup{hypertex=true,
	colorlinks=true,
	linkcolor=blue,
	anchorcolor=black,
	citecolor=green,
	linkbordercolor=green}
\usepackage{color} 
\usepackage{graphicx}
\usepackage{subfigure}
\usepackage{float}  
\usepackage{wrapfig} 

\def\tsc#1{\csdef{#1}{\textsc{\lowercase{#1}}\xspace}}
\tsc{WGM}
\tsc{QE}

\begin{document}
\let\WriteBookmarks\relax
\def\floatpagepagefraction{1}
\def\textpagefraction{.001}
\let\printorcid\relax % ORCID(s)

\shorttitle{Disentangled Representation via Variational AutoEncoder for Continuous Treatment Effect Estimation}   

\shortauthors{Ruijing Cui et al.}

\title[mode = title]{Disentangled Representation via Variational AutoEncoder for Continuous Treatment Effect Estimation}  

\author{Ruijing Cui}
\ead{cuiruijing@nudt.edu.cn}

\author{Jianbin Sun*}
\ead{sunjianbin@nudt.edu.cn}

\author{Bingyu He}
\ead{hebingyu20@nudt.edu.cn}

\author{Kewei Yang} 
\ead{kayyang27@nudt.edu.cn}

\author{Bingfeng Ge} 
\ead{bingfengge@nudt.edu.cn}

\address{College of Systems Engineering, National University of Defense Technology, Changsha 410073, P. R. China.}

\cortext[1]{Corresponding author}

\begin{abstract}
Continuous treatment effect estimation holds significant practical importance across various decision-making and assessment domains, such as healthcare and the military.  However, current methods for estimating dose-response curves hinge on balancing the entire representation by treating all covariates as confounding variables.  Although various approaches disentangle covariates into different factors for treatment effect estimation, they are confined to binary treatment settings.  Moreover, observational data are often tainted with non-causal noise information that is imperceptible to the human. Hence, in this paper, we propose a novel Dose-Response curve estimator via Variational AutoEncoder (DRVAE) disentangled covariates representation. Our model is dedicated to disentangling covariates into instrumental factors, confounding factors, adjustment factors, and external noise factors, thereby facilitating the estimation of treatment effects under continuous treatment settings by balancing the disentangled confounding factors.  Extensive results on synthetic and semi-synthetic datasets demonstrate that our model outperforms the current state-of-the-art methods.
\end{abstract}


\maketitle

\section{Introduction}
\label{Introduction}

Continuous treatment has emerged in many fields, such as healthcare, military, public policy, and economics. Due to the infeasibility of conducting randomized controlled trials, estimating the causal effects of continuous treatments to further obtain Average Dose-Response Function (ADRF) with the widespread accumulation of observational data while adjusting for confounding factors has become a crucial issue. The ADRF is crucial for decision-making. For example, in healthcare \cite{NatureHealthcare}, we need to estimate the effect of the specific dose of a drug in the recovery of a patient. In the military, conducting post-mission reviews is often necessary to inform the development of subsequent strategic plans. Specifically, it may be important to estimate the impact of certain tactical strategies or battlefield conditions on mission success rates. Taking Unmanned Aerial Vehicle (UAV) \cite{UAV} warfare as a running example, assessing the impact of flight altitude on reconnaissance probability may be meaningful for future similar military operations. The flight altitude is the specific dose of the treatment here and the reconnaissance probability is the potential outcome.

The primary obstacle in estimating treatment effects from observational data is known as the counterfactual outcomes and selection bias. In the setting of continuous treatment, only the factual outcome corresponding to a specific dosage is observed, while the outcomes that would have occurred if individuals had received different dosages are not available \cite{yao2021survey}. And each individual in the observed data is correlated to a specific treatment dosage, which may exhibit a dependency between the observed covariates and the treatment. The selection bias induced by that dependence can impact the estimation of the treatment effect. The methods for addressing the aforementioned two issues have benefited from cross-pollination and increasing interest in both causal inference and methine learning \cite{scholkopf2021toward} \cite{CMLsurvey}. Specifically, the core ideology of the recent methods \cite{BNN} \cite{TarnetCFRnet} \cite{DragonnetAndTR} \cite{DRNet} \cite{ADMIT} \cite{SurveyDCM} involves using encoder networks to project covariates into a unified representation space, which aims to balance the representations across treatment groups, thereby eliminating selection bias. Subsequently, it enables the estimation of counterfactual outcomes with these balanced representations.

Despite numerous previous attempts, two pivotal challenges persist. (i) A significant one is the limitation of the majority of existing studies to discrete treatment settings. And they cannot be easily extended to continuous settings. (ii) Another limitation is that most methods crudely balance the entire representation to reduce selection bias. The essence of this approach is treating all covariates as confounding variables; however, this introduces new estimation errors \cite{TEDVAE} \cite{DeRCFR}. Typically, covariates include instrumental factors, confounder factors, and adjustment factors \cite{DeRCFR}. The instrumental variables exert influence solely on the treatment. In contrast, confounder factors affect both the treatment and the outcome, while adjustment factors influence only the outcome. In this work, we also consider the presence of numerous external noise factors within the covariates, which are assumed to have no impact on either the treatment or the outcome. According to the backdoor criterion \cite{Pearl_2009}, we only need to balance the confounder factors in the representation space. Those factors implicit in covariates usually cannot be easily identified through manual recognition alone. In the running example, the dataset may encompasses UAV technical specifications, the proficiency levels of UAV operators, performance metrics of UAV payloads, environmental conditions, and other parameters deemed irrelevant to the flight. Empirical evidence and intuition suggest that the UAV technical specifications solely influence the flight altitude. The proficiency of operators is correlated with both the flight altitude and the probability of successful reconnaissance. Meanwhile, the performance parameters of the UAV payload only affect the reconnaissance probability. However, factors such as weather conditions and the type of reconnaissance target may also influence treatment and outcome, these significant battlefield environmental factors are difficult to be precisely identified.

Although some methods have attempted to address the first challenge, which is in the scope of this work, they still treated all covariates as confounder factors (i.e., the second challenge). Recent methods \cite{DRNet} \cite{TarnetCFRnet} discretize the continuous treatment, thereby training separate networks for each bin, which compromises the continuity \cite{ADRFcontinuity} of the ADRF. Other methods \cite{VCNet} \cite{GIKS} \cite{TransTEE} focus on designing sophisticated network architectures and loss functions to ensure the continuity of the ADRF while preventing high-dimensional covariates from overshadowing the treatment variable. These methods all assume that all covariates act as confounding factors and do not discriminate them during the network representation. Although many studies \cite{TEDVAE} \cite{DeRCFR} \cite{MIM-DRCFR} \cite{EDVAE} represent covariates as instrumental, confounding, and adjustment factors, they are all focused on binary treatments and cannot be readily extended to continuous settings. Besides, in practice, covariates frequently encompass a mix of discrete and continuous variables (e.g., as seen in commonly used benchmark datasets called IHDP \cite{IHDP} \cite{SurveyDCM}). In our running example, flight parameters typically include a substantial amount of boolean data to indicate the status of aircraft components. The prevailing approach does not adequately address this dichotomy when representing covariates.

\begin{figure}[h]
	\centering
	\includegraphics[width=0.6\textwidth]{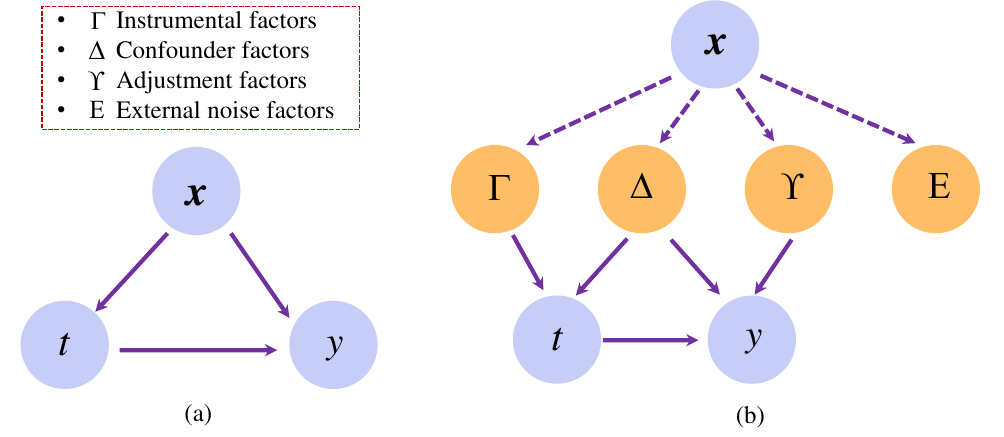}
	\caption{(a) The potential causal graph used by VCNet-liked methods (the varying coefficient structure proposed by \cite{VCNet} called VCNet). All covariates are treated as confounder factors here. (b) The latent causal graph of this work. Where dashed arrows signify representative relationships, while solid arrows indicate causal relationships.}
	\label{DAG}
\end{figure}

To address these challenges, a \textbf{D}ose-\textbf{R}esponse curve estimator via \textbf{V}ariational \textbf{A}uto\textbf{E}ncoder (DRVAE) is proposed through careful model design, which enables the disentangled representation for covariates to eliminate the selection bias and estimate counterfactuals under the continuous treatment setting. Contrary to VCNet-liked methods, which treat all covariates as confounding factors (Fig.\ref{DAG}(a)), we assume that covariates are determined by four latent factors: (1) Instrumental factors (2) Confounder factors (3) Adjustment factors, and (4) External noise factors (as shown in Fig.\ref{DAG}(b)). DRVAE first learns disentangled representations for each factor, thereby affording the opportunity to precisely estimate with only the relevant representations rather than the entire. The embedding from VAE enables our model to naturally estimate treatment effects under continuous settings. The model automatically produces continuous ADRF estimators (see Fig.\ref{curve} in Section \ref{Experiments}) while effectively preventing the treatment information from being lost in its high dimensional latent representation. Moreover, it discerns between continuous and discrete variables among covariates, which further enhances the precision of estimation. The principal contributions of this paper are as follows.

\begin{itemize}
	\item We propose the DRVAE model, which disentangles covariates into four distinct factors and learns balanced representations. Consequently, it enables the precision estimation for continuous treatment effects.
	\item We conduct extensive experiments on both synthetic and semi-synthetic datasets and show that we consistently outperform nine existing state-of-the-art methods. 
	\item We elucidate the reasons behind the observed gains by examining the presence of various factors. Simultaneously, we explore the performance of the model across four metrics, particularly in the context of increasing noise variables.
\end{itemize}

The subsequent sections are organized as follows: Section \ref{Previous work} provides a review of related work on disentangled representation and the ADRF estimation. Section \ref{Problem formulation} formulated the problem and the basic assumptions. In Section \ref{DRVAEsection}, we propose DRVAE and elucidate each component. Section \ref{Experiments} presents the experiments and corresponding outcomes. Ultimately, in Section \ref{Conclusion}, we summarize the paper and suggest potential future research directions.

\section{Previous work}
\label{Previous work}

\textbf{Disentangled representation for treatment effects estimation.} Learning disentangled representations is one of the problems of modern machine learning in the light of causal representation learning \cite{scholkopf2021toward}. The estimation of treatment effects is significantly enhanced by this perspective \cite{kuang2020causal}. Recent research has focused on disentangled representation learning for treatment effect estimation, aiming to reduce the influence of treatment-related factors and confounders on outcome prediction to mitigate selection bias \cite{chu2021learning}. Virtually most methods uniformly disentangled covariates into instrumental factors, confounding factors, and adjustment factors \cite{TEDVAE} \cite{FDVAE} \cite{DeRCFR} \cite{DR-CFR} \cite{EDVAE} \cite{VGANITE}. These methodologies exhibit several shared characteristics, particularly in the architecture of network structures and the formulation of loss functions. Collectively, they anticipate that instrumental and confounding factors within the representation space will facilitate the prediction of treatment. The confounding factors and adjustment factors, in conjunction with the treatment, are expected to predict the outcome. This anticipation is guided by the underlying common causal graph (see Fig.\ref{DAG}(b)). For instance, the Treatment Effect with Disentangled Autoencoder (TEDVAE) incorporates an auxiliary loss to articulate this predictive capability \cite{TEDVAE}. Conversely, the Decomposed Representations for CounterFactual Regression (DeR-CFR) \cite{DeRCFR} and Similarity preserved Individual Treatment Effect (SITE) \cite{SITE} models design specific prediction networks tailored for different treatments. To mitigate selection bias between treatment groups within the representation space, a prevalent strategy involves constraining the distributional distance between treatment and control groups \cite{DeRCFR} \cite{MIM-DRCFR}. This constraint aims to render the treatment and adjustment factors mutually independent, as dictated by the collider structure \cite{pearl2010} in the latent causal graph. Simultaneously, it also needs to ensure mutual independence among the disentangled representations. This mutual independence is critical to guarantee the accurate transference of information encapsulated within the covariates into their respective representation spaces. To achieve this end, two prevalent strategies are the utilization of orthogonal loss \cite{DeRCFR} and mutual information loss \cite{MIM-DRCFR} \cite{EDVAE}. In contrast to the neural network-based approaches discussed above, some methods adopt a data-driven strategy to directly decompose covariates \cite{DVD1} \cite{DVD2}. Furthermore, certain research focused on designing combinatorial optimization algorithms to discern and select diverse types of variables from covariates \cite{OAFP}.

However, these studies uniformly assume that the treatment is binary. The majority of network structures are predicated on this binary nature, necessitating the design of distinct networks for each treatment head. This foundational assumption significantly complicates the extension of these methods to continuous treatment settings.

\textbf{Estimating the average dose-response function (ADRF).} Recently, papers in ADRF estimation have utilized feed forward neural network for modeling. For its flexibility in modeling the complex causal relationship and potential for dealing with high-dimensional covariates \cite{masci2011stacked} \cite{BNN}. Several methods have been developed to estimate the ADRF within continuous settings. Building upon the Treatment-Agnostic Representation Network (TARNet) \cite{TarnetCFRnet} framework, Schwab et al. \cite{DRNet} introduced the Dose-Response Networks (DRNet), which discretizes continuous treatments and designs specific networks for each head.  However, this approach compromises the continuity of the ADRF curve. To address this challenge, Nie et al. \cite{VCNet} proposed the Varying Coefficient Networks (VCNet), which employs a varying coefficient prediction head rather than multiple heads, thereby preserving the continuity of the ADRF. VCNet also incorporates targeted regularization \cite{DragonnetAndTR} to achieve a doubly robust estimator across the entire ADRF curve, effectively enhancing the performance of finite samples. In conjunction with VCNet, Lokesh et al. \cite{GIKS} proposed gradient interpolation and kernel smoothing (GIKS) mode for treatment effect estimation, utilizing an auxiliary layer that infers counterfactuals based on data proximity to bridge the gap between training and testing distributions. Considering the diversity types of covariates and structures, Zhang et al. \cite{TransTEE} introduced the Transformer \cite{Transformer} as the treatment estimator (TransTEE), which also adapts to various types of treatments (i.e Binary, continuous, structural, and multiple).  Ioana et al. \cite{SCIGAN} proposed a model for estimating effects under continuous intervention conditions based on generative adversarial networks called SCIGAN. It is a counterfactual estimator extended from the GANITE \cite{GANITE} model in discrete intervention settings to continuous intervention settings.

Nevertheless, these methods primarily focus on the design of network structures and loss functions, without adequately distinguishing covariates.  That is, they simply treat covariates or their entire representations as confounding variables to adjust, which compromises the precision of treatment effect estimation. And they also fail to distinguish between discrete and continuous variables within the covariates, instead treating them uniformly.

Divergent from existing works, the DRVAE employs a variational autoencoder (VAE) \cite{VAE} to disentangle covariates into four distinct latent factors. Then we proceed to adjust the pertinent representations to mitigate selection biases, which enables us to achieve a robust estimation of continuous treatment effects.

\section{Problem formulation}
\label{Problem formulation}
We formalize the estimation problem of ADRF within the Rubin causal model framework \cite{rubin1974}, also known as the potential outcomes framework \cite{rubin2005}. Suppose we observe an \textit{i.i.d} sample $ \mathcal{D}: = \left\{ (\textbf{\textit{x}}_{ i } , t_{ i }, y_{ i })  \right\} _ { i = 1 } ^ { N }$ with size of which is $N$, where $(\textbf{\textit{x}}_{i},t_i,y_i)$ are independent realisations of random vector $(\textbf{\textit{X}},T,Y)$ with support $ (\mathcal{X},\mathcal{T},\mathcal{Y})$. Here $\textbf{\textit{X}}\in \mathbb{R}^{m}$ is referred as a vector of covariates, $T$ represents a continuous treatment within the interval $\mathcal{T}\in[0,1]$ in lines with \cite{VCNet}, and $Y$ signifies the outcome. Our goal is to estimate the \textit{Average Dose-Response Function (ADRF)} while eliminating selection bias in the continuous treatment setting. The ADRF is defined as
\begin{equation}
	\psi ( t ) : = \mathbb E ( Y | d o ( T = t ) ). 
\end{equation}
The $do(\cdot)$ operator signifies an intervention on a variable \cite{pearl2010}. The primary challenge is to learn $\psi ( t )$ from an observational dataset where each $\textbf{\textit{x}}$ is exposed to only one treatment dose, whose selection depends on $\textbf{\textit{x}}$ making the covariates correlated with the treatment. We assume that only a subset of information in the covariates is related to the treatment here. Our research is based on the following assumptions \cite{yao2021survey} \cite{imbens2015causal}. 

\textbf{Assumption 1} (Stable Unit Treatment Value Assumption (SUTVA)) There is no mutual influence among individuals, and each treatment exists in a singular version, wherein varying levels or doses of treatment are regarded as distinct treatments. Each individual's exposure to different treatments corresponds to different potential outcomes.

\textbf{Assumption 2} (Ignorability / Unconfoundedness) The potential outcome $Y(T = t)$ is independent of the treatment assignment given all covariates, i.e., $ Y ( T = t ) \bot t | \textbf{\textit{x}}.$

\textbf{Assumption 3} (Positivity / Overlap) Every unit should have non-zero probabilities to be assigned in each treatment group. Formally, $p( T = t | \textbf{\textit{X}} = \textbf{\textit{x}} ) > 0 , \forall t \in T , \forall \textbf{\textit{x}} \in \textbf{\textit{X}}$.

\textbf{Assumption 4} (Generation of Covariates) Given a set of covariates, denoted as $\textbf{\textit{x}}$, we assume $\textbf{\textit{x}}$ follows the joint distribution of instrumental factors $\Gamma$, confounder factors $\Delta$, adjustment factors $\Upsilon$ and other externally noise variables $\rm E$. Formally, $p(\textbf{\textit{x}}) = p(\Gamma,\Delta,\Upsilon,\rm E)$.

Assumptions 1 and 2 imply the identifiability of causal effects, wherein covariates $\textbf{\textit{x}}$ block all backdoor paths between treatment and outcome, allowing for the estimation of causal effects based on observational data.  Assumption 3 posits that irrespective of the covariates, each individual has the same opportunity to receive each treatment level, which is a standard assumption for constructing doubly robust estimators \cite{VCNet}.  Assumption 4, along with the latent causal graph (see Fig.\ref{DAG}(b)), indicates that only certain portions of information within the covariates serve as confounding factors, suggesting the potential for disentangling representations of covariates to attain more precise estimators.

\section{DRVAE: Our Proposed Approach}
\label{DRVAEsection}

\subsection{Disentanglement of Latent Representations}
\label{Disentanglement of Latent Representations}
The objective is to obtain unbiased estimates of the potential outcomes for a given set of input covariates: Based on the causal graph illustrated in Fig.\ref{DAG}(b), and guided by the back-door criterion \cite{Pearl_2009}, we can readily ascertain the specific form of the adjustment formula within the potential outcome framework \cite{rubin2005} as
\begin{equation}
	\begin{aligned}
		\psi ( t ) &= \mathbb E ( Y | d o ( T = t ) ) \\
		&= \mathbb{E}_{\textbf{\textit{X}}} \left[ \mathbb{E} ( Y \mid \textbf{\textit{X}} , T = t ) \right].
	\end{aligned}
\end{equation}
Noting that $\textbf{\textit{X}}$ is represented by $\Gamma,\Delta,\Upsilon$ and $\rm E$. So we obtain $ \pi ( t , \textbf{\textit{x}} ) : =\mathbb{E} ( Y \mid \textbf{\textit{X}} , T = t ) = \mathbb{E} ( Y \mid \Gamma,\Delta,\Upsilon,\rm E, T = t) $. Furthermore, we have $\Gamma,\rm E \bot \textit{Y}$ that yields 
$\psi ( t ) = \mathbb{E}_{\Delta,\Upsilon} \left[ \mathbb{E} ( Y \mid \Delta,\Upsilon , T = t ) \right]$, 
which enables us to plug in any machine learning model for $\psi ( t )$. Our contribution here is to propose the VAE based disentangled representation structure for estimate the $\hat{\pi} ( t , \textbf{\textit{x}} ) $ to further obtain $\psi ( t )$, which addresses difficulties confronting continuous treatment with confounding bias as discussed in the following paragraph.

\begin{figure}[h]
	\centering
	\includegraphics[width=0.8\textwidth]{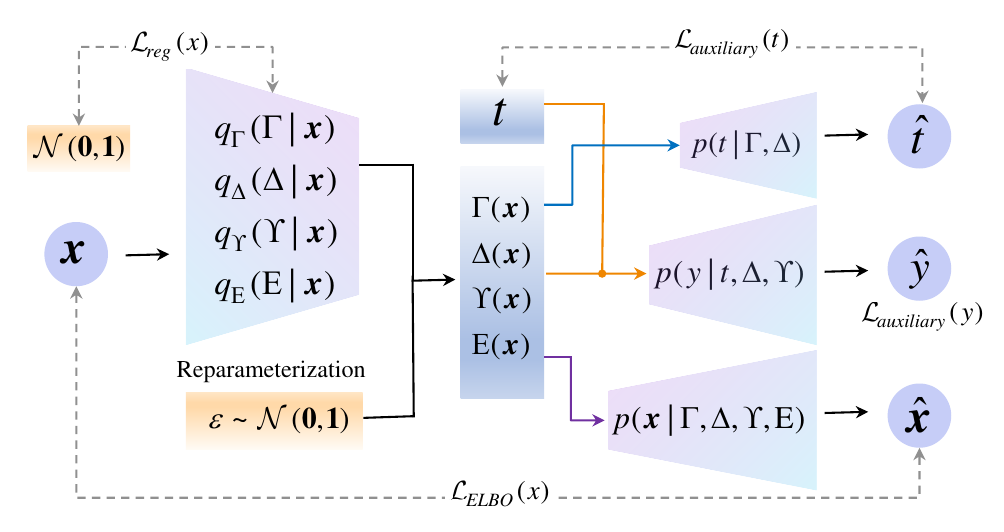}
	\caption{The framework of DRVAE.}
	\label{DRVAE}
\end{figure}

To predict $\pi ( t , \textbf{\textit{x}} ) $, an intuitive strategy involves training a neural network model using $(t, \textbf{\textit{x}})$ as input. Nevertheless, this methodology poses two primary challenges. Firstly, $t$ and $\textbf{\textit{x}}$ serve distinct roles in the causal graph, and the impact of $t$ on $y$ may be obscured by the high dimensionality of $\textbf{\textit{x}}$. Secondly, considering all variables in $\textbf{\textit{x}}$ as confounding factors introduce new biases in estimation. Previous works such as \cite{VCNet},\cite{DRNet}, and \cite{GIKS}, have indeed acknowledged the first issue. However, the second issue has been largely overlooked. DRVAE is crafted per the causal graph depicted in Fig.\ref{DAG}(b). Initially, it acquires disentangled representations of covariates, followed by the correction for selection bias utilizing pertinent representations rather than the entirety of them. The framework of DRVAE is shown in Fig.\ref{DRVAE}.

It is demonstrated that the impact of $t$ on $y$ can be discerned by recovering the confounding factors $\Delta$ from observational data. Our objective, therefore, is to infer the posterior distribution $p(\Gamma|\textbf{\textit{x}})$, $p(\Delta|\textbf{\textit{x}})$, $p(\Upsilon|\textbf{\textit{x}})$, $p(\rm E|\textbf{\textit{x}})$ for the four latent factors. Following the standard VAE design \cite{VAE} and the architecture of TEDVAE \cite{TEDVAE}, we resort to neural network-based variational inference to approximate the posterior $p _ { \theta } ( \textbf{\textit{x}} | \Gamma, \Delta, \Upsilon,\rm E) $. We employ four distinct encoders $ q_{\phi_{i}} (\textbf{\textit{z}}_{i}|\textbf{\textit{x}}), \textbf{\textit{z}}_{i}\in \{ \Gamma, \Delta, \Upsilon,\rm E\} $, which act as variational posteriors over the latent representations. And the prior distributions $ p(\Gamma), p(\Delta), p(\Upsilon),p(\rm E)$ are defined as
\begin{equation}
	p (\Gamma) = \prod _ { j = 1 } ^ { D _ {\Gamma} } \mathcal{N} (\Gamma_{j} | 0 , 1 ) ;  \\
	p (\Delta) = \prod _ { j = 1 } ^ { D _ {\Delta} } \mathcal{N} (\Delta_{j} | 0 , 1 ) ;  \\
	p (\Upsilon) = \prod _ { j = 1 } ^ { D _ {\Upsilon} } \mathcal{N} (\Upsilon_{j} | 0 , 1 ) ;  \\
	p ({\rm E}) = \prod _ { j = 1 } ^ { D _ {{\rm E}} } \mathcal{N} ({\rm E}_{j} | 0 , 1 ) ;  \\
\end{equation}
where $D$ are the parameters that determine the dimensions of instrumental, confounding, and external noise factors to infer from $\textbf{\textit{x}}$ and $j$ represent a certain dimension for relevant factors.
	
The generative model delineates the process of regenerating covariates $\textbf{\textit{x}}$, treatments $t$, and outcome $y$ based on the disentangled representations. Specifically, the generative models for $\textbf{\textit{x}},t,y$ are formulated as
\begin{equation}
\begin{aligned}
	&p ( \textbf{\textit{x}} |\Gamma, \Delta, \Upsilon,{\rm E}) = \prod _ { j = 1 } ^ { D_{\textbf{\textit{x}}} } p ( x _ { j } | \Gamma, \Delta, \Upsilon,{\rm E}) ,\\
	&p(t|\Gamma, \Delta) = \mathcal{N}(\mu=\widehat { \mu }_{t} , {\sigma} ^ { 2 } = \widehat { \sigma_{t} } ^ { 2 } ),\\
    &p ( y | t , \Delta,\Upsilon) =\mathcal{N}(\mu=\widehat { \mu }_{y} , {\sigma} ^ { 2 } = \widehat { \sigma_{y} } ^ { 2 } ).
\end{aligned}
\end{equation}
Where the conditional distributions of $t$ and $y$ are conformed as Gaussian distributions due to their continuous nature. As mentioned earlier, the covariates $\textbf{\textit{x}}$ consist of both binary and continuous features. We model them separately as Bernoulli distributions and Gaussian distributions, specifically, we have
\begin{equation}
\begin{aligned}
 p(x^{bin}_j|\Gamma, \Delta, \Upsilon,{\rm E}) &= Bern(\sigma(f_{\theta_{bin}}(\Gamma, \Delta, \Upsilon,{\rm E}))),\\
 p(x^{con}_j|\Gamma, \Delta, \Upsilon,{\rm E}) &= \mathcal{N}(\mu=\widehat { \mu }_{x_{j}} , {\sigma} ^ { 2 } = \widehat { \sigma}_{x_{j}} ^ { 2 } ).
\end{aligned}
\end{equation}
$\sigma(\cdot)$ is the logistic function. And $\mu$ and $\sigma$ represent the mean and standard deviation for the Gaussian distribution, respectively. They are parametrized by the relevant representations as
\begin{equation}
\begin{aligned}
&\widehat { \mu }_{x_{j}} = f_{\theta_{x_{j}}}(\Gamma, \Delta, \Upsilon,{\rm E}),  \widehat { \sigma}_{x_{j}} ^ { 2 }  = f_{\theta_{x_{j}}}(\Gamma, \Delta, \Upsilon,{\rm E}),\\
&\widehat { \mu }_{t} =f_{\theta_{\mu_{t}}}(\Gamma, \Delta) ,  \widehat { \sigma_{t} } ^ { 2 } = f_{\theta_{\sigma_{t}}}(\Gamma, \Delta),\\
&\widehat { \mu }_{y} =f_{\theta_{\mu_{y}}}(t,\Delta, \Upsilon) ,  \widehat { \sigma_{y} } ^ { 2 } = f_{\theta_{\sigma_{y}}}(t,\Delta, \Upsilon), \\
\end{aligned}
\end{equation}
where all $f$ are neural networks parametrized by their own parameters $\theta_{(\cdot)}$.

In the inference model, the variational approximations of the posteriors are defined as:
\begin{equation}
q _ { \phi _ { i} } ( \textbf{\textit{z}} _ {i } | \textbf{\textit{x }}) = \prod _ { j = 1 } ^ { D _ { \textbf{\textit{z}} _ { i } } } \mathcal{N} ( \mu = \widehat { \mu } _ {\textit{\textbf{z}}_{i}} , \sigma ^ { 2 } = \widehat { \sigma } _ { \textit{\textbf{z}}_{i} } ^ { 2 } ), \textbf{\textit{z}}_{i}\in \{ \Gamma, \Delta, \Upsilon,\rm E\}. 
\end{equation}
These Gaussian distributions serve as approximations of the posteriors for the instrumental, confounding, adjustment, and external noise latent variables. They are parameterized by neural networks, enabling us to learn the distribution parameters from the data.

\subsection{Training}
\label{Training}
The process of disentangling covariates into latent representations follows the standard setup of VAE, where the relevant parameters can be optimized by maximizing the evidence lower bound (ELBO) \cite{ELBO}. Specifically, 
\begin{equation}
\mathcal{L} _ { E L B O } ( \textbf{\textit{x}} ) = \alpha \mathbb{E}_{ q _ { \phi_{\Gamma}} q _ { \phi_{\Delta}} q _ { \phi_{\Upsilon}}q _ { \phi_{{\rm E}}} } \left[ \log p _ { \theta } ( \textbf{\textit{x}} |\Gamma, \Delta, \Upsilon,{\rm E} ) \right]  - \beta \sum_{i} D _ { K L } ( q _ { \phi _ { i} } ( \textbf{\textit{z}} _ {i } | \textbf{\textit{x}} ) | | p _ { \theta _ { i } } ( \textbf{\textit{z}} _ { i } ) ).
\end{equation}

Where $D _ { K L }(\cdot||\cdot)$ represents the KL divergence between the two distributions. Previous studies frequently estimated treatment probability by utilizing the complete covariate representation \cite{DragonnetAndTR} \cite{VCNet}. However, this method may inadvertently incorporate extraneous information, including adjustment factors and external noise factors, from the representation. Contrary to this, as illustrated by the causal graph Fig.\ref{DAG}(b), the disentangled representations of $\Gamma$ and $\Delta$ enable the prediction of $t$. Furthermore, $\Delta$, $\Upsilon$, and $t$ are required to produce $y$ as the outcome. Simultaneously, we discard the representation $\rm E$, which is assumed to contain information about external noise data. The prediction loss is defined as
\begin{equation}
	\begin{aligned}
	\mathcal{L} _ { auxiliary} ( t, y ) &=\gamma \mathcal{L} _ { auxiliary} ( t ) + \delta \mathcal{L} _ { auxiliary} ( y )\\
	&= \gamma \mathbb{E}_{ q _ { \phi_{\Gamma}} q _ { \phi_{\Delta}}} \left[ \log p _ { \theta } (t |\Gamma, \Delta) \right]  +\delta \mathbb{E}_{ q _ { \phi_{\Delta}} q _ { \phi_{\Upsilon}}} \left[ \log p _ { \theta } (y |t, \Delta,\Upsilon ) \right].
	\end{aligned}
\end{equation}

To ensure that the latent variable distribution aligns with the anticipated prior distribution throughout the model training process, thereby enhancing the generalization capability and mitigating overfitting, we incorporate a regularization loss
\begin{equation}
		\mathcal{L} _ {reg} (\textbf{\textit{x}}) = \sum_{i} \lambda \mathbb{E}_{q_{ \phi_{i}}} \left[\log \frac{p _ { \theta_{i}}(z_{i})}{q _ { \phi _ { i} } ( z _ {i } | \textbf{\textit{x }}) }  \right].
\end{equation}

Finally, the DRVAE can be trained by maximizing the objective of follows
\begin{equation}
	\mathcal{L} _ {DRVAE} (\textbf{\textit{x}}, t, y ) = \mathcal{L} _ { E L B O } ( \textbf{\textit{x}} ) +\mathcal{L} _ {auxiliary} (t, y ) +\mathcal{L} _ {reg} (\textbf{\textit{x}}).
\end{equation}
The hyperparameters $\alpha, \beta, \gamma, \delta$ and $\lambda$ balance the different objectives. During the inference of counterfactuals, we employ encoders $q_ { \phi _ {\Delta} } (\Delta | \textbf{\textit{x }}) $ and $q _ { \phi _ {\Upsilon} } (\Upsilon | \textbf{\textit{x }}) $ to derive the posterior distributions for the confounding and adjustment factors. Subsequently, we perform $l$ iterations of sampling and employ the decoder $p ( y | t , \Delta,\Upsilon)$ to compute the average outcome $y$. Specifically, $\hat{y}_{i}(t) = \hat{\pi} ( t , \textbf{\textit{x}}_{i} )  = \widehat { \mu }_{i,y}$. So, given $n$ test instances, $\hat{\psi} ( t ) = \mathbb{E}_{\Delta,\Upsilon} \left[ \mathbb{E} ( Y \mid \Delta,\Upsilon , T = t ) \right] =  \frac { 1 } { n } \sum _ { i = 1 } ^ { n } \hat{y}_{i}(t)$. Where $t$ represents the value after the intervention with the $do(\cdot)$ operator in the test set.

\section{Experiments}
\label{Experiments}
\textbf{Dataset.} The inherent unobservability of counterfactual outcomes renders it impractical to directly collect such data from real-world observations. Consequently, synthetic and semi-synthetic datasets are commonly utilized to evaluate the empirical performance of models, particularly within the setting of continuous treatment effect estimation. Synthetic datasets explicitly integrate the data-generating mechanisms, i.e., the causal relationships between variables, facilitating the straightforward derivation of the ground truth counterfactuals. In alignment with \cite{VCNet} \cite{TransTEE}, our study utilizes one set of synthetic data and two sets of commonly used semi-synthetic data: IHDP \cite{IHDP} \cite{IHDP2} and News. 

To emulate potential external sources of noise inherent in real-world data, we introduce two types of noise variables into the synthetic datasets, which are irrelevant to both the treatment and outcome. These noise variables are sampled from Normal distributions $\varepsilon_{con}\!\sim\!\mathcal{N}(2, 10)$, and Bernoulli distributions $\varepsilon_{bin} \!\sim\! Bern(p)$ with $p\!\sim\!\text{Unif}[0,1]$, respectively. Depending on the dimensions of the introduced noise, the synthetic datasets are labeled as Simu($k$), where $k={1,2,3,4,5}$. IHDP originated from the Infant Health Development Program and was initially utilized for binary treatment effect estimation. Treatments in the dataset were allocated via a randomized experiment. It encompasses 747 subjects with 25 covariates. To estimate the continuous treatment effect, the dataset underwent first modifications in \cite{VCNet}, wherein synthetic treatments and outcomes were assigned. The News dataset comprises 3000 randomly sampled news items from The New York Times corpus, which was originally introduced as a benchmark for binary setting \cite{BNN} and adapted in \cite{SCIGAN} for continuous treatment. We utilized the data version provided by VCNet \cite{VCNet}.

The fundamental details of all datasets are outlined in Table \ref{3Dataset}. A comprehensive description of the data generation process can be found in the \nameref{Appendix}
\begin{table}[H]
	\centering
	\caption{The fundamental details of the utilized datasets. Taking “6+5+10” as an example, where “6” denotes the number of covariates in the original data, “5” represents the count of continuous noise variables $\varepsilon_{con}$, and “10” signifies the number of binary noise variables $\varepsilon_{bin}$.}
	\label{3Dataset}
	\begin{tabular}{l|ccccccc}
		\toprule 
		Size& Simu($1$)  & Simu($2$)   & Simu($3$)   & Simu($4$)   & Simu($5$)   & IHDP & News \\ 
		\midrule
		covariates   & 6+5+10 & 6+10+20 & 6+15+30 & 6+20+40 & 6+25+50 & 25   & 498  \\
		observations & 700    & 700     & 700     & 700     & 700     & 747  & 3000 \\ 
		\bottomrule 
	\end{tabular}
\end{table}

\textbf{Baselines and Settings.} Considering our research scope, we have selected several state-of-the-art (SOTA) methods that are most relevant to our work for comparison. The following are the specific details.

\begin{itemize}
	\item TARNet and its targeted regularization \cite{DragonnetAndTR} version TARNet\_TR. It was proposed by Shalit et al \cite{TarnetCFRnet}. for the estimation of binary treatments. Its network architecture predominantly features a shared network alongside treatment-specific head networks. While preserving the structural integrity of their network, Patrick et al \cite{DRNet}. introduced dosage as an additional input with the aim of leveraging it as an estimator for continuous treatment effects.
	\item DRNet and its targeted regularization version DRNet\_TR. The DRNet \cite{DRNet} discretizes the continuous treatment into multiple intervals and allocates a dedicated head network to each interval for estimating the dose-response curve. Its remarkable aspect lies in its consideration of heterogeneity between treatment and covariates. More precisely, within each hidden layer of the network architecture, treatment inputs are merged with covariates to prevent high-dimensional covariates from overwhelming the treatment.
	\item VCNet and its targeted regularization version VCNet\_TR. VCNet \cite{VCNet} introduces a varying coefficient structure for estimating outcomes in the setting of continuous treatment. The varying coefficient model \cite{Varying-coefficient1} \cite{Varying-coefficient1} allows the weights of the prediction head to be continuous functions of the treatment, thereby facilitating VCNet to maintain the continuity of the dose-response curve by employing continuous activation functions. VCNet is commonly used as a benchmark method for estimating continuous treatment effects.
	\item TransTEE. Zhang et al. \cite{TransTEE} employ Transformer Backbones to estimate treatment effects. The attention mechanisms are utilized to govern interactions between treatments and covariates, leveraging the structural similarities in potential outcomes to control for confounding variables. TransTEE demonstrates adaptability to diverse types of treatments and covariates, encompassing discrete, continuous, tabular, and textual.
	\item GIKS. Lokesh et al. \cite{GIKS} proposed the GIKS method, which utilizes Gradient Interpolation and Kernel Smoothing to estimate continuous treatment effects.  Specifically, GIKS first employs Gradient Interpolation for close-to-observed treatments and then downweighs the high variance inferences with a Gaussian Process based Kernel Smoothing to estimate the counterfactual outcomes.
\end{itemize}

To ensure a fair and reliable comparison, the parameters of the SOTA methods are set to their optimal values as reported in their original papers. For the TARNet, DRNet, VCNet, and their corresponding TR versions models, we adopt the implementations provided by \cite{VCNet}. Implementations for TransTEE and GIKS were sourced directly from the respective authors. For DRVAE, both the encoder and decoder networks are optimized with the Adam optimizer, with the network architecture being fully connected and utilizing ELU activation functions, characterized by an alpha parameter set to 1.0. The hyperparameters $\alpha, \beta, \gamma, \delta$ and $\lambda$ within the objective function are meticulously calibrated in $\{0.1,0.5,1.0,10\}$, and the learning rate and weight decay rate are aligned with the configurations in \cite{TEDVAE}, with values fixed at 1e-3 and 1e-4, respectively. During the prediction of $y$ with the decoder, sampling of both $\Delta$ and $\Upsilon$ is repeated $l=20$ times. To mitigate the potential for treatment information to be overshadowed by high-dimensional disentangled representations, we have specified the dimensions of the four latent factors to be 1. Actually, ablation studies have indicated that the dimensionality of these latent factors exerts a negligible influence on the outcome. For the purpose of hyperparameter optimization, we adhered to the dataset configurations as outlined in \cite{VCNet}. The optimal hyperparameter configurations for each dataset are presented in Table \ref{Parameters}.

Hardware configuration: Intel i7-13700K 3.40-GHz CPU and 32 GB of memory. Software configuration: Python with Pytorch 2.2.2.
\begin{table}[!H]
	\caption{Parameters setting for DRVAE. The Hidden-dim indicates the dimensions of the hidden layers, while Num-layers specify the number of such layers.}
	\label{Parameters}
	\begin{tabular}{l|ccccccccccc}
		\toprule 
		Dataset& Hidden-dim & Num-layers & \multicolumn{1}{c}{Epochs} & Batch size & Learning rate&Weight decay & $\alpha$ & $\beta$ & $\gamma$ & $\delta$ & $\lambda$ \\
		 \midrule
		Simu($k$) & 128    & 3     & 100     & 64     & 1e-3&1e-4      & 0.1    & 1.0   & 1.0 & 0.1   & 1.0       \\
		IHDP    & 100        & 3          & 100   & 64   & 1e-3&1e-4      & 0.1    & 1.0   & 1.0 & 1.0   & 1.0      \\
		News    & 100        & 3          & 80   & 256   & 1e-3&1e-4     & 0.1    & 1.0   & 1.0 & 1.0   & 1.0     \\ 
		\bottomrule
	\end{tabular}
\end{table}

\textbf{Evaluation Metrics.} Given $n$ test instances, we employ one primary metric and three auxiliary metrics to assess the performance of the model in estimating the continuous treatment effects. The metrics utilized cover several desirable aspects of model performance, including dose-response curves at both the population and individual levels, as well as identifying the optimal dosage points for individuals. Following \cite{VCNet}, we utilize the average mean squared error (AMSE) on the test set to evaluate the performance of models in estimating the population level dose-response curves. AMSE is defined as
\begin{equation}
AMSE=\int_{\mathcal{T}}[\frac{1}{n}\sum_{i=1}^{n}(\hat{y}_{i}(t)-y_{i}(t))]^{2}p(t)dt.
\end{equation}
Since the integral of $t$ is intractable, we approximate the AMSE by applying all the $t$ values present in the current dataset to each individual, yielding  $A\hat{M}SE=\frac{1}{|\mathcal{T}|}\sum_{t\in\mathcal{T}}[\frac{1}{n}\sum_{i=1}^{n}(\hat{y}_{i}(t)-y_{i}(t))]^{2}$. Where $|\mathcal{T}|$ is the number of different treatment values.

To ascertain the extent to which the model captures the entire range of individual dose-response curves, we employ the mean integrated square error (MISE) as used by \cite{DRNet}. The metric measures the discrepancy between the actual dose-response $y$ and the predicted dose-response $\hat{y}$, as estimated by the model across $n$ individuals and the entire range of specific dosages. MISE is presented as
\begin{equation}
		MISE=\frac{1}{n}\sum_{i=1}^{n}[\int_{\mathcal{T}}(\hat{y}_i(t)-y_i(t))^2dt].
\end{equation}
Following \cite{DRNet}, we used Romberg integration with 64 equally spaced samples from $y_i(t)$ and $\hat{y}_i(t)$ to compute the inner integral over the range of dosage parameters necessary for the MISE metric. Simultaneously, within this integration step size, we compute the individual-level mean square error ($i$-MSE) as
\begin{equation}
i-MSE=\frac{1}{n}\sum_{i=1}^{n}[\frac{1}{|\mathcal{T}|}\sum_{t\in \mathcal{T}}(\hat{y}_i(t)-y_i(t))^2]. 
\end{equation}

To evaluate the performance of the model in calculating the optimal dosage point for the corresponding treatment at the individual level, we calculate dosage policy error (DPE) \cite{DRNet}. DPE is outlined as
\begin{equation}
 DPE = \frac { 1 } { n } \sum _ { i = 1 } ^ { n } ( y _ { i } ( t ^ { * } ) - y _ { i } ( \hat { t }^ { * } ) ) ^ { 2 },
\end{equation}
where $t^*=\underset{t\in \mathcal{T}}{\arg\max}y_{i}(t)$, $\hat{t}^*=\underset{t\in \mathcal{T}}{\arg\max}\hat{y}_{i}(t)$. The DPE measures the discrepancy between the two true outcomes $y _ { i } ( t ^ { * } )$ and  $y _ { i } ( \hat { t }^ { * } )$, which corresponds to the true optimal dosage point $ t ^ { * }$ and the model-estimated optimal dosage point $ \hat { t }^ { * } $, respectively.

\subsection{Comparison with SOTA Methods}
\label{Comparison with SOTA Methods}
We report the mean and standard deviation of the aforementioned four metrics across 10 repetitions, each initialized with a distinct random seed.  In consistent with prior works \cite{SCIGAN} \cite{DRNet}, we report the square root transformations of the MISE and DPE. A smaller metric value means the better performance of the model.
\begin{table}[H]
	\centering
	\caption{Performance comparison between DRVAE and baselines on Simu($k$) with four metrics.}
	\label{4MetricsFORSimu}
	\resizebox{0.8495\linewidth}{!}
	{
	\begin{tabular}{l|lllll}
		\Xhline{0.8pt}  
		\multicolumn{1}{l|}{\multirow{2}{*}{Method}} & \multicolumn{5}{c}{AMSE (Mean$\pm$Standard deviation)}                                                                                                                            \\ \cline{2-6} 
		\multicolumn{1}{l|}{} & \multicolumn{1}{c}{Simu(1)} & \multicolumn{1}{c}{Simu(2)} & \multicolumn{1}{c}{Simu(3)} & \multicolumn{1}{c}{Simu(4)} & \multicolumn{1}{c}{Simu(5)} \\ 
		\hline
		Tarnet               & 0.0736$\pm$0.0229           & 0.0867$\pm$0.0201           & 0.1098$\pm$0.0174           & 0.1203$\pm$0.0422           & 0.1238$\pm$0.0458  \\
		Tarnet\_tr           & >$10^3$     & >$10^3$      & >$10^3$      & >$10^3$      & >$10^3$      \\
		Drnet                & 0.0660$\pm$0.0113            & 0.0981$\pm$0.0441           & 0.1101$\pm$0.0321           & 0.1215$\pm$0.0320            & 0.1661$\pm$0.0700  \\
		Drnet\_tr            & >$10^2$       & >$10^2$      & >$10^2$      & >$10^2$   & >$10^2$    \\
		Vcnet                & 0.0250$\pm$0.0107            & 0.0475$\pm$0.0215           & 0.0418$\pm$0.0182           & 0.0618$\pm$0.0623           & 0.0500$\pm$0.0154  \\
		Vcnet\_tr            & 0.0325$\pm$0.0201           & 0.0468$\pm$0.0183           & 0.0480$\pm$0.0225            & 0.0413$\pm$0.0202           & 0.0831$\pm$0.0561  \\
		TransTEE             & 0.2036$\pm$0.0615           & 0.3510$\pm$0.0808            & 0.3420$\pm$0.0947            & 0.2934$\pm$0.0525           & 0.3569$\pm$0.1024 \\
		TransTEE\_tr         & 0.1473$\pm$0.0603           & 0.3779$\pm$0.0705           & 0.3739$\pm$0.0792           & 0.3609$\pm$0.0616           & 0.3208$\pm$0.0607  \\
		GIKS                 & 0.0251$\pm$0.0119           & 0.0517$\pm$0.0449           & 0.0312$\pm$0.0132           & 0.0453$\pm$0.0325           & 0.0316$\pm$0.0189  \\ 
		\hline
		DRVAE (\textbf{ours}) & \textbf{0.0147$\pm$0.0075}  &\textbf{ 0.0209$\pm$0.0080}  & \textbf{0.0183$\pm$0.0094}  & \textbf{0.0181$\pm$0.0089 }  & \textbf{0.0193$\pm$0.0090}  \\
		\hline\hline
		\multicolumn{1}{l|}{\multirow{2}{*}{Method}} & \multicolumn{5}{c}{$\sqrt{\rm MISE}$ (Mean$\pm$Standard deviation)}                                                                                                                            \\ \cline{2-6} 
		\multicolumn{1}{l|}{} & \multicolumn{1}{c}{Simu(1)} & \multicolumn{1}{c}{Simu(2)} & \multicolumn{1}{c}{Simu(3)} & \multicolumn{1}{c}{Simu(4)} & \multicolumn{1}{c}{Simu(5)} \\ 
		\hline
		Tarnet               & 0.8450$\pm$0.0664            & 0.8833$\pm$0.0710            & 0.9088$\pm$0.0326           & 0.9574$\pm$0.0345           & 0.9559$\pm$0.0488  \\
		Tarnet\_tr           & >$10^1$          & >$10^1$         & >$10^1$          & >$10^1$          & >$10^1$         \\
		Drnet                & 0.8625$\pm$0.0361           & 0.9076$\pm$0.0433           & 0.9077$\pm$0.0365           & 0.9508$\pm$0.0380            & 0.9642$\pm$0.0266 \\
		Drnet\_tr            & >$10^1$           & >$10^1$          & >$10^1$         & >$10^1$         & >$10^1$        \\
		Vcnet                & 0.7472$\pm$0.0165           & 0.8565$\pm$0.0618           & 0.8791$\pm$0.0222           & 0.8949$\pm$0.0547           & 0.8898$\pm$0.0349  \\
		Vcnet\_tr            & 0.8050$\pm$0.0319            & 0.9132$\pm$0.0382           & 0.9058$\pm$0.0529           & 0.8877$\pm$0.0327           & 0.9078$\pm$0.0437  \\
		TransTEE             & 0.7940$\pm$0.0980             & 1.1135$\pm$0.3697           & 1.0312$\pm$0.2471           & 1.0205$\pm$0.3096           & 1.0689$\pm$0.1841\\
		TransTEE\_tr         & 0.8181$\pm$0.1432           & 0.9757$\pm$0.1373           & 1.0358$\pm$0.2963           & 1.0858$\pm$0.3434           & 0.9094$\pm$0.1745  \\
		GIKS                 & 0.6874$\pm$0.0379           & 0.7379$\pm$0.0563           & 0.7189$\pm$0.0514           & 0.7508$\pm$0.0590            & 0.7675$\pm$0.0378  \\ \hline
		DRVAE (\textbf{ours}) & \textbf{0.6134$\pm$0.0091}  & \textbf{0.6133$\pm$0.0152} & \textbf{0.6146$\pm$0.0179}& \textbf{0.6164$\pm$0.0148} & \textbf{0.6184$\pm$0.0078}  \\ 
		\hline\hline
		\multicolumn{1}{l|}{\multirow{2}{*}{Method}} & \multicolumn{5}{c}{$\sqrt{\rm DPE}$ (Mean$\pm$Standard deviation)}                                                                                                                             \\ \cline{2-6} 
		\multicolumn{1}{l|}{}& \multicolumn{1}{c}{Simu(1)} & \multicolumn{1}{c}{Simu(2)} & \multicolumn{1}{c}{Simu(3)} & \multicolumn{1}{c}{Simu(4)} & \multicolumn{1}{c}{Simu(5)} \\ 
		\hline
		Tarnet               & 0.9934$\pm$0.0525           & 1.0469$\pm$0.0612           & 1.0848$\pm$0.0606           & 1.1731$\pm$0.0991           & 1.1313$\pm$0.0747  \\
		Tarnet\_tr           & 1.2466$\pm$0.4071           & 1.2060$\pm$0.2248            & 1.1203$\pm$0.1372           & 1.4054$\pm$0.2538           & 1.2107$\pm$0.1431  \\
		Drnet                & 1.0632$\pm$0.0648           & 1.1144$\pm$0.0907           & 1.0912$\pm$0.1082           & 1.1400$\pm$0.0879             & 1.1624$\pm$0.1237 \\
		Drnet\_tr            & 1.5762$\pm$0.5022           & 1.1461$\pm$0.1281           & 1.3155$\pm$0.2119           & 1.3778$\pm$0.3183           & 1.1761$\pm$0.1431 \\
		Vcnet                & 0.9395$\pm$0.0434           & 1.0349$\pm$0.0791           & 1.0679$\pm$0.0467           & 1.1140$\pm$0.1359            & 1.1115$\pm$0.0956  \\
		Vcnet\_tr            & 0.9582$\pm$0.0524           & 1.0025$\pm$0.0541           & 1.0953$\pm$0.1114           & 1.0316$\pm$0.0522           & 1.1849$\pm$0.2242 \\
		TransTEE             & 1.0263$\pm$0.2531           & 1.0177$\pm$0.1779           & 1.1686$\pm$0.3612           & 1.0545$\pm$0.3542           & 1.1176$\pm$0.2947  \\
		TransTEE\_tr         & 1.0348$\pm$0.3473           & 1.0256$\pm$0.1670            & 1.0126$\pm$0.1298           & 0.9310$\pm$0.0533            & 0.9949$\pm$0.1465  \\
		GIKS                 & 0.9383$\pm$0.0565           & 0.9651$\pm$0.0764   & \textbf{0.8945}$\pm$0.0324    & 0.9598$\pm$0.0666           & 0.9586$\pm$0.0515 \\ \hline
		DRVAE (\textbf{ours})   & \textbf{0.9063$\pm$0.0423} & \textbf{0.8690$\pm$0.0422}  & 0.9133$\pm$\textbf{0.0280} & \textbf{0.9132$\pm$0.0460} & \textbf{0.8827$\pm$0.0341} \\ \hline\hline
		\multicolumn{1}{l|}{\multirow{2}{*}{Method}}& \multicolumn{5}{c}{$i$-MSE (Mean$\pm$Standard deviation)}                                                                                                                             \\ \cline{2-6} 
		\multicolumn{1}{l|}{}& \multicolumn{1}{c}{Simu(1)} & \multicolumn{1}{c}{Simu(2)} & \multicolumn{1}{c}{Simu(3)} & \multicolumn{1}{c}{Simu(4)} & \multicolumn{1}{c}{Simu(5)} \\ 
		\hline
		Tarnet               & 0.8028$\pm$0.1237           & 0.8777$\pm$0.1349           & 0.9243$\pm$0.0749           & 1.0175$\pm$0.0765           & 1.0120$\pm$0.1100\\
		Tarnet\_tr           & >$10^3$    & >$10^3$  & >$10^3$    & >$10^3$    & >$10^3$       \\
		Drnet                & 0.8349$\pm$0.0697           & 0.9269$\pm$0.0996           & 0.9231$\pm$0.0846           & 1.0071$\pm$0.0793           & 1.0322$\pm$0.0569\\
		Drnet\_tr            & >$10^3$    & >$10^3$    & >$10^3$    & >$10^3$  & >$10^3$    \\
		Vcnet                & 0.6308$\pm$0.0313           & 0.8501$\pm$0.1362           & 0.8865$\pm$0.0458           & 0.9437$\pm$0.1391           & 0.9171$\pm$0.0758  \\
		Vcnet\_tr            & 0.7363$\pm$0.0666           & 0.9773$\pm$0.0882           & 0.9605$\pm$0.1366           & 0.9140$\pm$0.0793            & 0.9604$\pm$0.1107 \\
		TransTEE             & 0.6969$\pm$0.1621           & 1.4755$\pm$1.0503           & 1.2208$\pm$0.5619           & 1.2124$\pm$0.7369           & 1.2587$\pm$0.4179  \\
		TransTEE\_tr         & 0.7441$\pm$0.2762           & 1.0864$\pm$0.3057           & 1.2447$\pm$0.7423           & 1.3849$\pm$0.8864           & 0.9391$\pm$0.3534 \\
		GIKS                 & 0.5529$\pm$0.0737           & 0.6580$\pm$0.1121            & 0.6844$\pm$0.1982           & 0.7550$\pm$0.1977            & 1.0256$\pm$0.3543  \\ \hline
		DRVAE (\textbf{ours}) & \textbf{0.4172$\pm$0.0159} & \textbf{0.4191$\pm$0.0248} & \textbf{0.4244$\pm$0.0300}  &\textbf{ 0.4238$\pm$0.0255} & \textbf{0.4244$\pm$0.0128}   \\ 
		\Xhline{0.8pt}
	\end{tabular}
}
\end{table}

Table \ref{4MetricsFORSimu} presents a comparison of the DRVAE with all state-of-the-art continuous treatment effect estimation methods across the datasets Simu($k$) with four metrics. Bold indicates the best results. We make the following observations: (1) The DRVAE outperforms all baseline methods across four metrics and on five Simu datasets, demonstrating the effectiveness in estimating continuous treatment effects. It also highlights the rationality of its model architecture and loss function design. Although the $\sqrt{\rm DPE}$ metric on the Simu(3) dataset did not surpass GIKS, DRVAE still ranked second with significant competitiveness among all methods. (2) The TR versions of TARNet and DRNet significantly distinguished themselves from other methods across the AMSE, $\sqrt{\rm MISE}$, and $i$-MSE metrics. A metric value exceeding 10 indicates a possible failure of convergence in the network. However, the TR versions of VCNet and TransTEE yielded normal results. TARNet and DRNet discretize continuous treatments in their network design, differing from all other methods. Hence, designing networks to discretize continuous treatments and coupling them with TR may struggle to adapt to various data types, even though they perform adequately in individual metrics such as $\sqrt{\rm DPE}$. (3) In terms of a single dataset, DRVAE exhibits better stability compared to other methods, as evidenced by the smallest standard deviation across repeated experiments among all methods. It indicates that DRVAE is less affected by noise factors in the data generation process, accurately preserving causal information between variables while encoding non-causal information into appropriate external noise factors.

\begin{figure}
	\centering
	\subfigure[AMSE]
	{
		\label{fig:subfig:a}
		\includegraphics[width=0.35\linewidth]{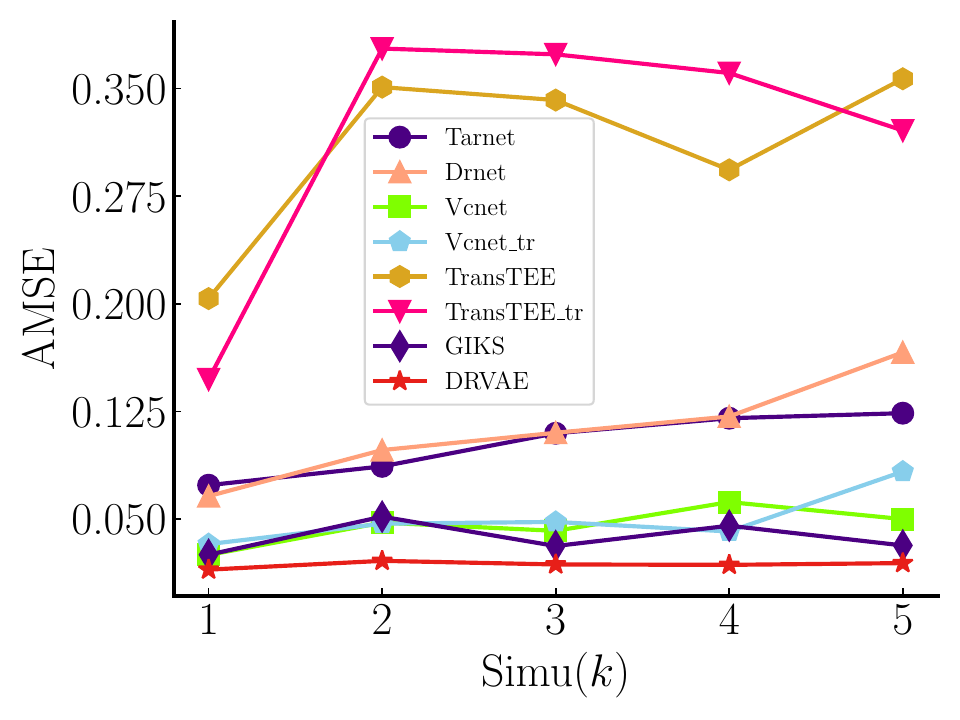}
	}
	\hspace{0.00001\linewidth}
	\subfigure[$\sqrt{\rm MISE}$]
	{
		\label{fig:subfig:b}
		\includegraphics[width=0.35\linewidth]{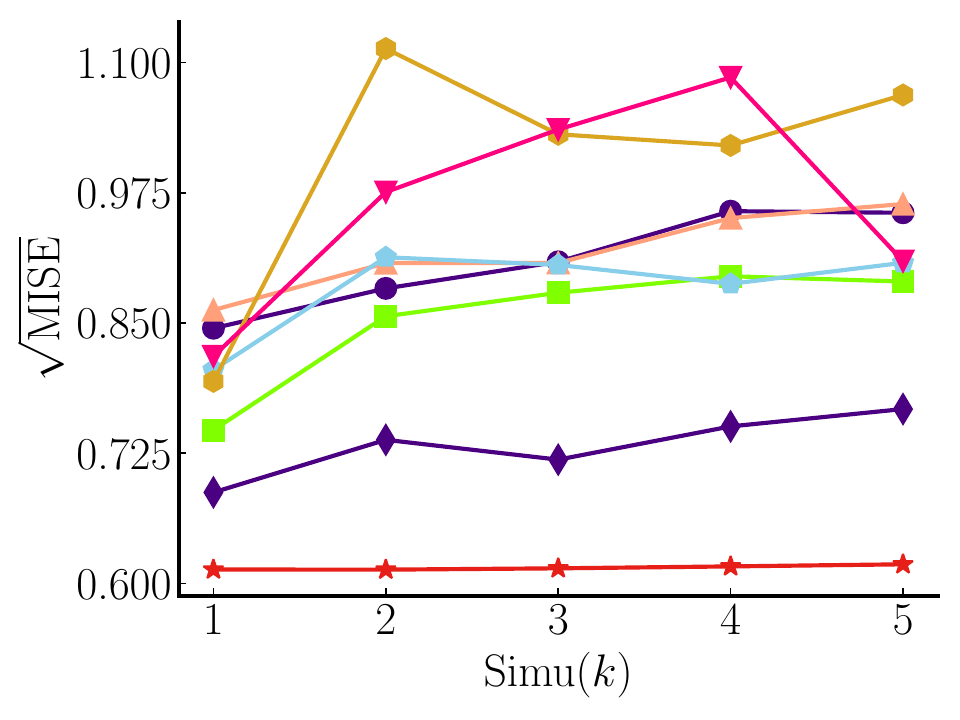}
	}
	\vfill
	\subfigure[$\sqrt{\rm DPE}$]
	{
		\label{fig:subfig:c}
		\includegraphics[width=0.35\linewidth]{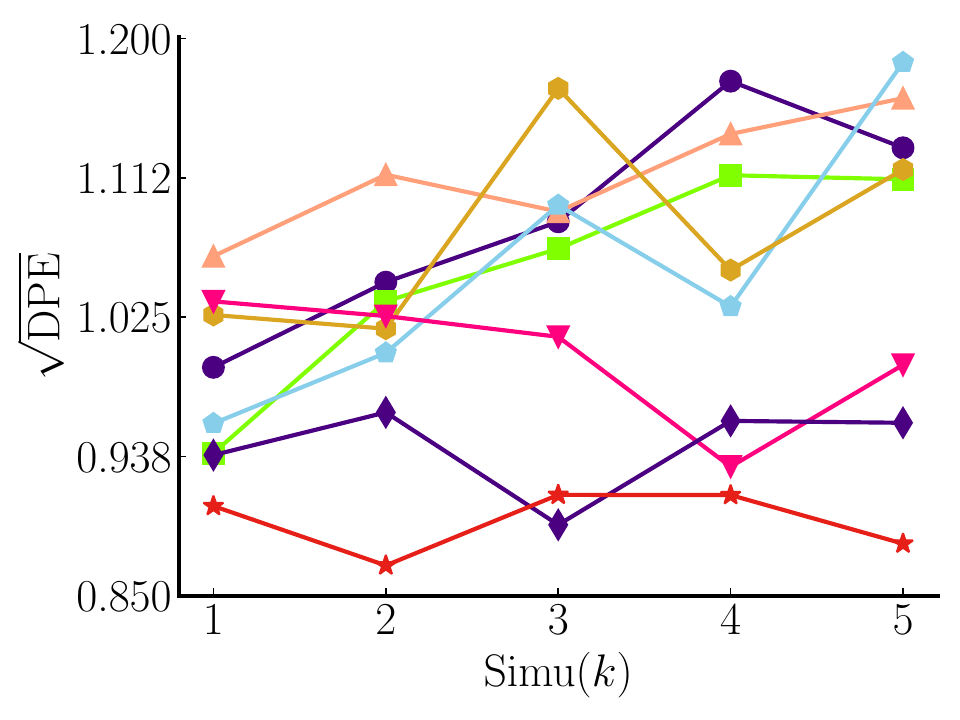}
	}
	\hspace{0.00001\linewidth}
	\subfigure[$i$-MSE]
	{
		\label{fig:subfig:d}
		\includegraphics[width=0.35\linewidth]{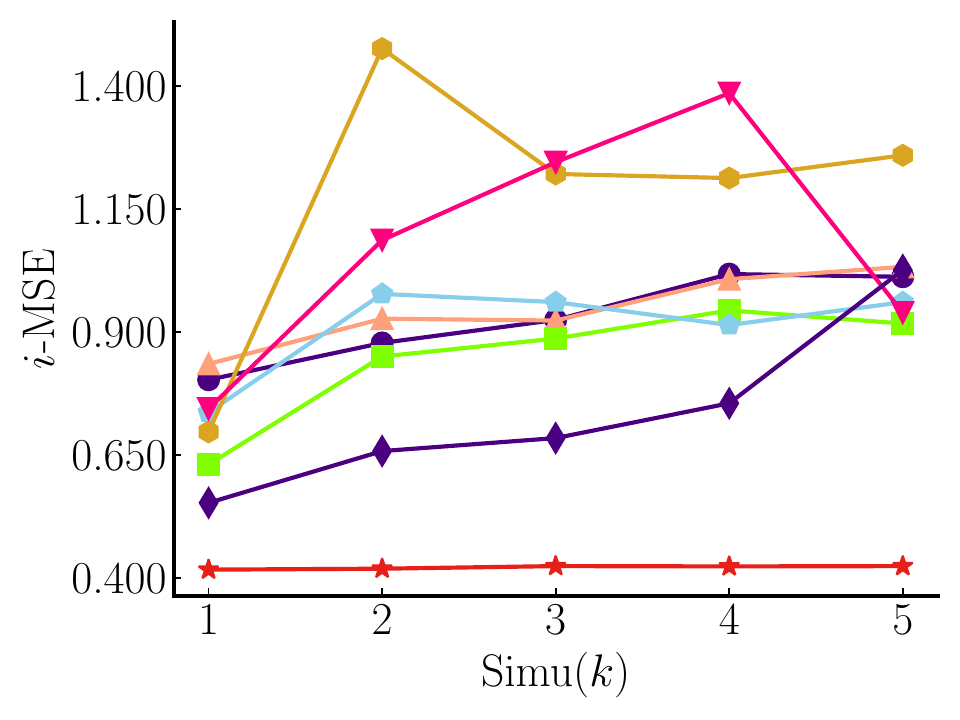}
	}
	\caption{ A comparative analysis of the performance trends of the DRVAE against other benchmark methods on varying Simu($k$) datasets. The increase in external noise factors does not significantly impact the performance of our model.}
	\label{plot}
\end{figure}

Figure \ref{plot} delineates the performance trajectories of diverse models across four metrics, with respect to the escalating presence of exogenous noise in the dataset. It is evident that the DRVAE remains impervious to the addition of non-causal noise, adeptly discerning the underlying true causal mechanism from the data. Specifically, datasets Simu(1) through Simu(5) are progressively infused with incremental noise, as detailed in Table \ref{3Dataset}, which all are initially treated as covariates. The previous works fail to parse the covariates' composition, treating them uniformly as confounding factors. In contrast, the DRVAE meticulously disentangles partial covariate information as confounders, while deeming other partial elements as latent noise, thereby enhancing model versatility across datasets. As depicted in Figure \ref{plot}, despite the surge in noise factors, the DRVAE sustains optimal performance on metrics such as the AMSE, $\sqrt{\rm MISE}$, and $i$-MSE, outperforming baselines without significant fluctuation. The TransTEE and its TR version manifest more pronounced oscillations, whereas other methods exhibit a modest decline in precision. This variability stems from the noise interference with the authentic causal information. Analogously, barring the TransTEE\_tr, other methods on the $\sqrt{\rm DPE}$ follow an erratic descending pattern, whereas the DRVAE maintains a relatively stable course. Even when the original data incorporates a mix of binary and continuous noise variables, the DRVAE achieves superior results. In conclusion, the DRVAE adeptly extracts and retains covariate information with genuine causal ties to both treatment and outcome variables, discarding extraneous non-causal inputs, and thereby ensuring precise estimation of continuous treatment effects.

As elaborated in the Appendix, the semi-synthetic IHDP dataset has 25 covariates, which can be divided into 3 groups: $S_{con} = {1, 2, 3, 5, 6}$, $S_{dis,1} = {4, 7 \sim 15}$, and $S_{dis,2} = {16 \sim 25}$. $S_{con}$ serves as confounding variables that influences both $t$ and $y$, $S_{dis,1}$ is the adjustment variables for it influences only $y$, while $S_{dis,1}$ act as instrumental variables affecting only $t$. It can be seen that external noise factors E are not present in the synthetic IHDP dataset.  Hence, the DRVAE model is configured with a dimensionality of $D_{\rm E}=0$, while maintaining the dimensions of the remaining three latent factors at 1. The results of 10 repeated experiments are shown in Table \ref{IHDP}. Compared to other baseline methods, DRVAE achieved the best performance across all four metrics. Notably, there is a significant improvement in the AMSE metric for the DRVAE method. A meticulous examination of the dataset indicates that this superior performance is likely due to the substantial number of binary covariates in the IHDP, which were not distinctly accounted for by previous methods.  During the training of DRVAE, we reconstructed these using a Bernoulli distribution, thereby increasing the precision of estimation.  On the other hand, as shown in the appendix, the data generation process involves sampling a certain amount of noise from a Gaussian distribution.  DRVAE enhances the adaptability to noise of the model through a disentangling strategy.
\begin{table}[H]
	\centering
	\caption{Performance comparison between DRVAE and baselines on IHDP with four metrics.}
	\label{IHDP}
	\begin{tabular}{l|llll}
		\toprule 
		Method & \multicolumn{1}{c}{AMSE} & \multicolumn{1}{c}{$\sqrt{\rm MISE}$} & \multicolumn{1}{c}{$\sqrt{\rm DPE}$} & \multicolumn{1}{c}{$i$-MSE} \\ 
		\midrule 
		Tarnet       & 1.2456$\pm$0.6570        & 1.5421$\pm$0.1251        & 2.9801$\pm$1.7552       & 3.1945$\pm$0.5360        \\
		Tarnet\_tr   & 0.8098$\pm$0.2402       & 1.4598$\pm$0.1294        & 1.6583$\pm$1.3913       & 2.7861$\pm$0.4447       \\
		Drnet        & 1.1911$\pm$0.3901       & 1.5608$\pm$0.0837        & 2.5890$\pm$1.6672        & 3.2613$\pm$0.3194       \\
		Drnet\_tr    & 0.8191$\pm$0.3171       & 1.4382$\pm$0.0840         & 1.7313$\pm$1.3377       & 2.6999$\pm$0.3006       \\
		Vcnet        & 0.5070$\pm$0.2825        & 1.2352$\pm$0.0829        & 0.6842$\pm$0.1177       & 2.0919$\pm$0.2904       \\   
		Vcnet\_tr    & 0.3902$\pm$0.3754       & 1.2140$\pm$0.2191         & 1.1603$\pm$1.5248       & 1.9957$\pm$0.6175       \\
		TransTEE     & 2.8987$\pm$0.4502       & 2.4349$\pm$0.4707        & 3.0730$\pm$2.0533        & 7.3144$\pm$2.3465       \\
		TransTEE\_tr & 2.9406$\pm$0.5689       & 2.4508$\pm$0.2600          & 4.6541$\pm$1.0375       & 7.3023$\pm$1.3769       \\
		GIKS         & 0.4709$\pm$0.1405       & 2.6015$\pm$0.1748        & 0.6196$\pm$0.0527       & 7.9061$\pm$1.0753       \\
		\midrule 
		DRVAE (\textbf{ours})        & \textbf{0.0945$\pm$0.0436}       & \textbf{1.1694$\pm$0.0339}         & \textbf{0.5529$\pm$0.0541 }      & \textbf{1.9188$\pm$0.1893 }      \\
		\bottomrule 
	\end{tabular}
\end{table}

As depicted in Table \ref{News}, for the News dataset, DRVAE continues to outperform other baseline methods on the primary metric, AMSE. However, all methods exhibit significant deviations on the other three metrics. As analyzed in \cite{GIKS}, the training instances of the News dataset possess an unconstrained additive Gaussian noise component sampled from $\mathcal{N}(0,0.5)$ in their outcome $y$, as shown in Eq. \ref{Enews}. This mechanism does not affect other datasets since their $y$ values are considerably larger than 0.5, but for News, it poses a substantial impact on $y$. Nonetheless, DRVAE demonstrates superior performance over other methods in terms of AMSE.
\begin{table}[]
	\centering
	\caption{Performance comparison between DRVAE and baselines on News with four metrics.}
	\label{News}
	\begin{tabular}{l|llll}
		\toprule 
		Method & \multicolumn{1}{c}{AMSE} & \multicolumn{1}{c}{$\sqrt{\rm MISE}$} & \multicolumn{1}{c}{$\sqrt{\rm DPE}$} & \multicolumn{1}{c}{$i$-MSE} \\ 
		\midrule 
		Tarnet       & 0.0557$\pm$0.0044       & 1.1300$\pm$0.0087          & 1.1969$\pm$0.0180        & 1.6005$\pm$0.0280        \\
		Tarnet\_tr   & 0.0345$\pm$0.0051       & 1.1241$\pm$0.0093        & 1.2052$\pm$0.0331       & 1.5904$\pm$0.0268       \\
		Drnet        & 0.0557$\pm$0.0044       & 1.1302$\pm$0.0102        & 1.2044$\pm$0.0206       & 1.6069$\pm$0.0301       \\
		Drnet\_tr    & 0.0343$\pm$0.0054       & 1.1256$\pm$0.0090         & 1.1986$\pm$0.0292       & 1.5924$\pm$0.0281       \\
		Vcnet        & 0.0492$\pm$0.0406       & 1.0751$\pm$0.0263        & 1.0664$\pm$0.0307       & 1.3461$\pm$0.0659       \\
		Vcnet\_tr    & 0.0310$\pm$0.0371        & 1.0996$\pm$0.0209        & 1.0764$\pm$0.0581       & 1.4055$\pm$0.0617       \\
		TransTEE     & 0.7901$\pm$0.1108       & 1.6737$\pm$0.2940         & 1.1575$\pm$0.0495       & 3.2325$\pm$1.0154       \\
		TransTEE\_tr & 0.8062$\pm$0.1786       & 1.6192$\pm$0.2923        & 1.1488$\pm$0.0704       & 3.0408$\pm$1.0818       \\
		GIKS         & 0.0413$\pm$0.0311       & 1.0423$\pm$0.0339        & 1.1981$\pm$0.0690        & 1.3105$\pm$0.1118       \\
		\midrule 
		DRVAE (\textbf{ours})        & \textbf{0.0155$\pm$0.0086}       & 1.1190$\pm$0.0079         & 1.1969$\pm$0.0259       & 1.5827$\pm$0.0268       \\ 
		\bottomrule 
	\end{tabular}
\end{table}

\begin{figure}
	\centering
	\subfigure[Simu($1$)]
	{
		\label{fig:subfig:a}
		\includegraphics[width=0.31\linewidth]{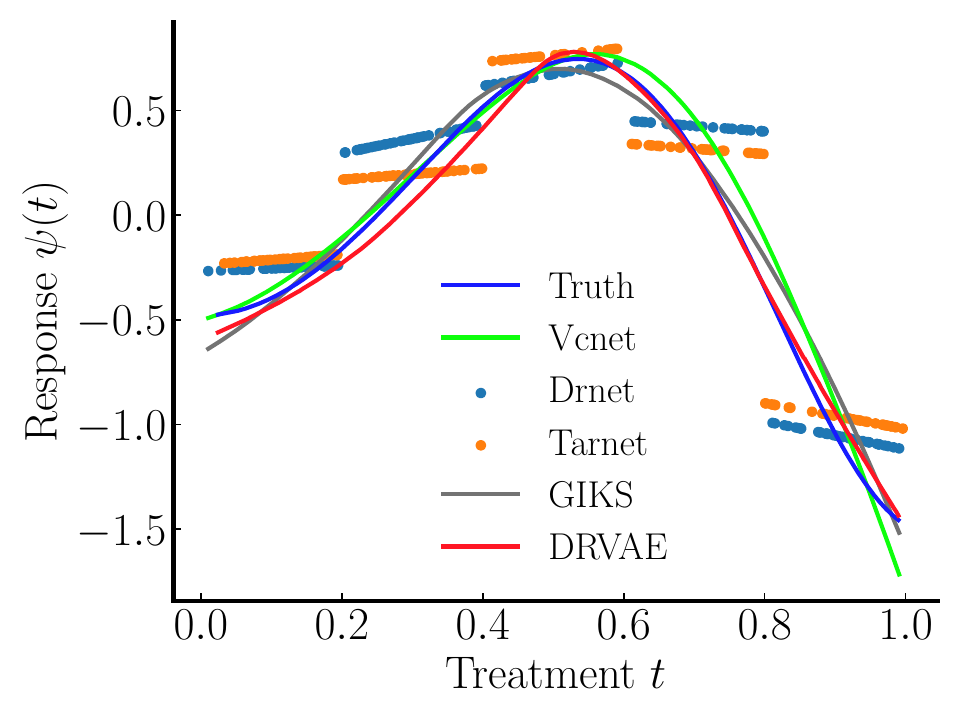}
	}
	\hspace{0.00001\linewidth}
	\subfigure[IHDP]
	{
		\label{fig:subfig:b}
		\includegraphics[width=0.31\linewidth]{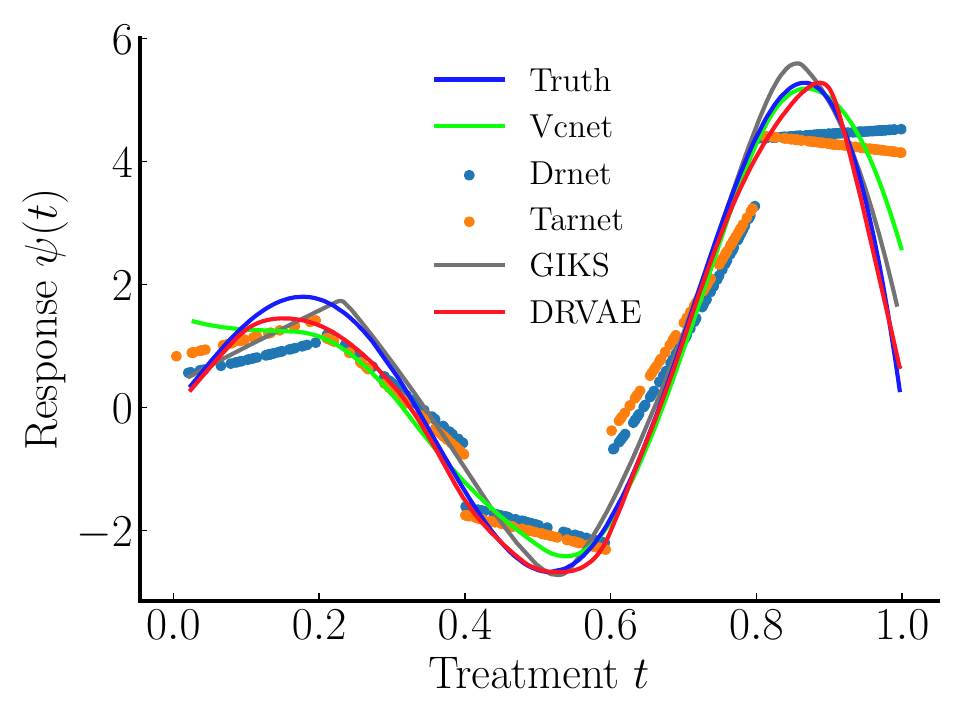}
	}
	\hspace{0.00001\linewidth}
	\subfigure[News]
	{
		\label{fig:subfig:c}
		\includegraphics[width=0.31\linewidth]{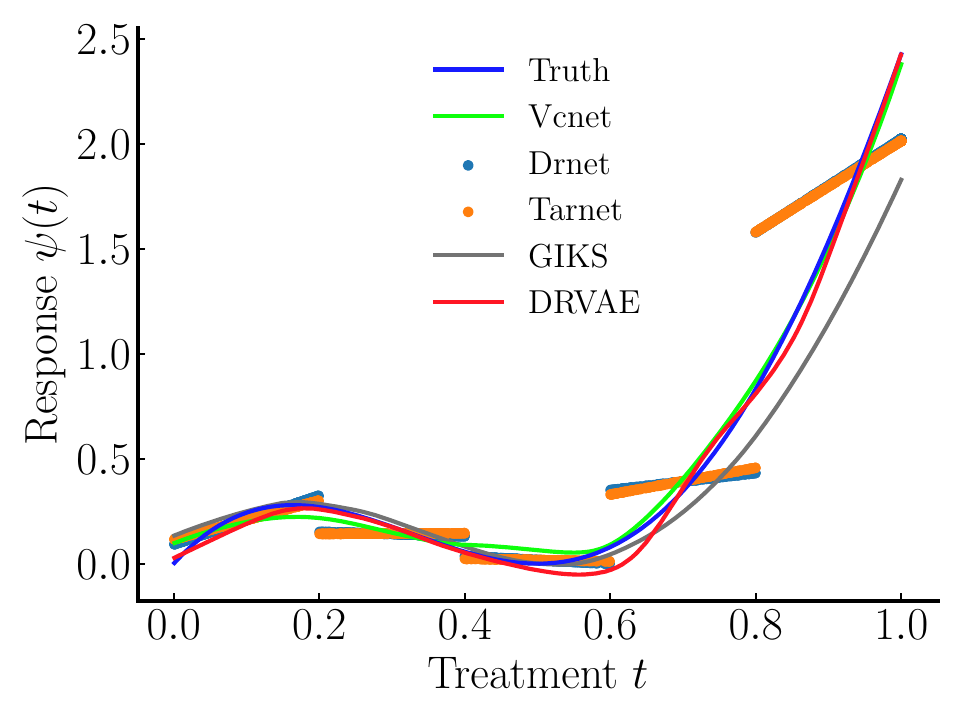}
	}
	\caption{Estimated ADRF on the testing set from a typical run of DRVAE and other five baseline methods.}
	\label{curve}
\end{figure}

As illustrated in Fig. \ref{curve}, we also qualitatively observed the performance of DRVAE in estimating the ADRF curve. The TR versions of some methods were not visualized in the figure due to a lack of convergence. It can be discerned that the DRNet and TARNet methods disrupted the continuity of the ADRF curve, as noted by VCNet \cite{VCNet}. In contrast, the DRVAE model not only effectively maintains the continuity of the curve but also achieves more accurate estimation. Since DRVAE outperforms all other methods on AMSE across all datasets, the curves seem coincident.

\subsection{Ablation study}
\label{Ablation study}
The section examines the impact of disentangling the representation of four latent factors and the configuration of the objective function on estimating continuous treatment effects.

\textbf{Impact of Alternative Factors.} Here, we delve into the influence of the presence or absence of four key factors and their dimensionality on the outcomes. The findings detailed in Section \ref{Comparison with SOTA Methods} are predicated on the dimensions of these four factors being unified at  $D_{\textbf{\textit{z}}_{i}} = 1,\textbf{\textit{z}}_{i} \in \{\Gamma, \Delta, \Upsilon,\rm E\}$, a decision made to mitigate the risk of high-dimensional latent factors overshadowing the causal effect of $t$ on $y$.

\begin{wrapfigure}{l}{5cm}
	\centering
	\includegraphics[width=0.30\textwidth]{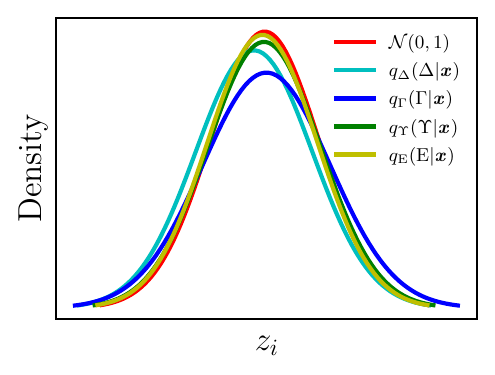}
	\caption{Latent factors distribution.}
	\label{Gaussian}
\end{wrapfigure} 
Table \ref{absence} delineates the performance metrics of the DRVAE when applied to the Simu(1) dataset under various scenarios of latent factor absence. Specifically, the absence of a factor is indicated by setting its corresponding dimension to 0; for example, the notation “0-1-1-1” signifies that the absence of $\Gamma$, while the other three factors retain their standard dimensions of 1. Where “Baseline*” refers to the best outcomes achieved by all comparative methods across the respective datasets, as derived from Table \ref{4MetricsFORSimu}. It is noteworthy that even in the absence of certain latent factors, the DRVAE consistently surpasses other benchmark methods across a range of metrics. Intriguingly, the omission of specific latent factors does not markedly affect the performance. This robustness is attributed to the disentangled representations, which are all governed by the same underlying distribution. Indeed, due to the rigorous training regimen, the four types of latent factors are closely aligned with the standard Gaussian distribution, as illustrated in Fig. \ref{Gaussian}. Consequently, when estimating the distribution of $p ( y | t , \Delta,\Upsilon) $, the sampled values of $\Delta$ and $\Upsilon$ exhibit a high degree of similarity.

\begin{table}[H]
	\centering
	\caption{The performance of DRVAE on the dataset Simu($1$) under different absence of the factors.}
	\label{absence}
	\begin{tabular}{c|cccc}
		\toprule 
		$D_{\Gamma}$-$D_{\Delta}$-$D_{\Upsilon}$-$D_{\rm E}$& \multicolumn{1}{c}{AMSE} & \multicolumn{1}{c}{$\sqrt{\rm MISE}$} & \multicolumn{1}{c}{$\sqrt{\rm DPE}$} & \multicolumn{1}{c}{$i$-MSE} \\ 
		\midrule 
		\rowcolor{gray!25}Baseline* & 0.0250$\pm$0.0107        & 0.6874$\pm$0.0379        & 0.9383$\pm$0.0565       & 0.5529$\pm$0.0737       \\
		1-1-1-1   & 0.0147$\pm$0.0075       & 0.6134$\pm$0.0091        & 0.9063$\pm$0.0423       & 0.4172$\pm$0.0159       \\
		0-1-1-1   & 0.0143$\pm$0.0079       & 0.6146$\pm$0.0131        & 0.8852$\pm$0.0257       & 0.4192$\pm$0.0188       \\
		1-0-1-1   & 0.0154$\pm$0.0077       & 0.6123$\pm$0.0128        & 0.9034$\pm$0.0397       & 0.4181$\pm$0.0190        \\
		1-1-0-1   & 0.0150$\pm$0.0085        & 0.6129$\pm$0.0116        & 0.8787$\pm$0.0436       & 0.4175$\pm$0.0169       \\
		1-1-1-0   & 0.0150$\pm$0.0072        & 0.6095$\pm$0.0117        & 0.8831$\pm$0.0380        & 0.4134$\pm$0.0187       \\ 
		\bottomrule
	\end{tabular}
\end{table}

\begin{table}[H]
\centering
\caption{The performance of DRVAE on the dataset Simu($k$) under different dimensions of the factors.}
\label{SIMU}
\begin{tabular}{l|ccccc}
\Xhline{0.8pt}  
\multirow{2}{*}{$D_{\Gamma}$-$D_{\Delta}$-$D_{\Upsilon}$-$D_{\rm E}$}                      & \multicolumn{5}{c}{AMSE (Mean$\pm$Standard deviation)}                                                                                                                  \\ \cline{2-6} 
                                       & \multicolumn{1}{c}{Simu(1)} & \multicolumn{1}{c}{Simu(2)} & \multicolumn{1}{c}{Simu(3)} & \multicolumn{1}{c}{Simu(4)} & \multicolumn{1}{c}{Simu(5)} \\ \hline
\rowcolor{gray!25} Baseline*                              & 0.0250$\pm$0.0107          & 0.0468$\pm$0.0183         & 0.0312$\pm$0.0132         & 0.0413$\pm$0.0202         & 0.1238$\pm$0.0458         \\
5-5-5-5                                & 0.0142$\pm$0.0068         & 0.0236$\pm$0.0110          & 0.0180$\pm$0.0095          & 0.0214$\pm$0.0096         & 0.0202$\pm$0.0082         \\
10-10-10-10                            & 0.0167$\pm$0.0079         & 0.0256$\pm$0.0105         & 0.0200$\pm$0.0091           & 0.0203$\pm$0.0105         & 0.0210$\pm$0.0081          \\
15-15-15-15                            & 0.0159$\pm$0.0078         & 0.0250$\pm$0.0090           & 0.0214$\pm$0.0107         & 0.0199$\pm$0.0098         & 0.0207$\pm$0.0074         \\
20-20-20-20                            & 0.0164$\pm$0.0076         & 0.0264$\pm$0.0113         & 0.0218$\pm$0.0101         & 0.0220$\pm$0.0107          & 0.0206$\pm$0.0073         \\
50-50-50-50                            & 0.0164$\pm$0.0076         & 0.0287$\pm$0.0111         & 0.0232$\pm$0.0104         & 0.0228$\pm$0.0115         & 0.0222$\pm$0.0079         \\ \hline\hline
\multirow{2}{*}{$D_{\Gamma}$-$D_{\Delta}$-$D_{\Upsilon}$-$D_{\rm E}$}                      & \multicolumn{5}{c}{$\sqrt{\rm MISE}$ (Mean$\pm$Standard deviation)}                                                                                                                  \\ \cline{2-6} 
                                       & Simu(1)                     & Simu(2)                     & Simu(3)                     & Simu(4)                     & Simu(5)                     \\  \hline
\rowcolor{gray!25} Baseline*                              & 0.6874$\pm$0.0379         & 0.7379$\pm$0.0563         & 0.7189$\pm$0.0514         & 0.7508$\pm$0.0590          & 0.7675$\pm$0.0378         \\
5-5-5-5                                & 0.6109$\pm$0.0114         & 0.6156$\pm$0.0175         & 0.6160$\pm$0.0153          & 0.6209$\pm$0.0181         & 0.6196$\pm$0.0099         \\
10-10-10-10                            & 0.6137$\pm$0.0109         & 0.6158$\pm$0.0159         & 0.6133$\pm$0.0146          & 0.6171$\pm$0.0183         & 0.6171$\pm$0.0103         \\
15-15-15-15                            & 0.6159$\pm$0.0097         & 0.6183$\pm$0.0164         & 0.6149$\pm$0.0160          & 0.6170$\pm$0.0175         & 0.6182$\pm$0.0070          \\
20-20-20-20                            & 0.6123$\pm$0.0120         & 0.6164$\pm$0.0173         & 0.6166$\pm$0.0144          & 0.6192$\pm$0.0165         & 0.6187$\pm$0.0094         \\
50-50-50-50                            & 0.6108$\pm$0.0106         & 0.6170$\pm$0.0162         & 0.6179$\pm$0.0162          & 0.6176$\pm$0.0185         & 0.6185$\pm$0.0108         \\ \hline\hline
\multirow{2}{*}{$D_{\Gamma}$-$D_{\Delta}$-$D_{\Upsilon}$-$D_{\rm E}$}                      & \multicolumn{5}{c}{$\sqrt{\rm DPE}$ (Mean$\pm$Standard deviation)}                                                                                                                   \\ \cline{2-6} 
                                       & Simu(1)                     & Simu(2)                     & Simu(3)                     & Simu(4)                     & Simu(5)                     \\  \hline
\rowcolor{gray!25} Baseline*                              & 0.9383$\pm$0.0565         & 0.9651$\pm$0.0764         &  0.8945$\pm$0.0324         & 0.9310$\pm$0.0533          & 0.9586$\pm$0.0515         \\
5-5-5-5                                & 0.9101$\pm$0.0291         & 0.9155$\pm$0.0544         & 0.9280$\pm$0.0415          & 0.9060$\pm$0.0570         & 0.8966$\pm$0.0427         \\
10-10-10-10                            & 0.8845$\pm$0.0272         & 0.8895$\pm$0.0462         & 0.9065$\pm$0.0314          & 0.9085$\pm$0.0377         & 0.8809$\pm$0.0296         \\
15-15-15-15                            & 0.8914$\pm$0.0248         & 0.9205$\pm$0.0363         & 0.9041$\pm$0.0355          & 0.9025$\pm$0.0367         & 0.8824$\pm$0.0532         \\
20-20-20-20                            & 0.9044$\pm$0.0327         & 0.9062$\pm$0.0418         & 0.9040$\pm$0.0359          & 0.9051$\pm$0.0353         & 0.8936$\pm$0.0497         \\
50-50-50-50                            & 0.8961$\pm$0.0366         & 0.8956$\pm$0.0291         & 0.9150$\pm$0.0352          & 0.9067$\pm$0.0483         & 0.8915$\pm$0.0298         \\ \hline\hline
\multirow{2}{*}{$D_{\Gamma}$-$D_{\Delta}$-$D_{\Upsilon}$-$D_{\rm E}$}                      & \multicolumn{5}{c}{$i$-MSE (Mean$\pm$Standard deviation)}                                                                                                                   \\ \cline{2-6} 
                                       & Simu(1)                     & Simu(2)                     & Simu(3)                     & Simu(4)                     & Simu(5)                     \\  \hline
\rowcolor{gray!25} Baseline*                              & 0.5529$\pm$0.0737         & 0.6580$\pm$0.1121          & 0.6844$\pm$0.1982         & 0.7550$\pm$0.1977          & 0.9171$\pm$0.0758         \\
5-5-5-5                                & 0.4152$\pm$0.0173         & 0.4231$\pm$0.0275         & 0.4266$\pm$0.0271         & 0.4309$\pm$0.0298         & 0.4261$\pm$0.0180          \\
10-10-10-10                            & 0.4196$\pm$0.0163         & 0.4235$\pm$0.0262         & 0.4232$\pm$0.0249         & 0.4253$\pm$0.0306         & 0.4264$\pm$0.0162         \\
15-15-15-15                            & 0.4209$\pm$0.0166         & 0.4257$\pm$0.0271         & 0.4249$\pm$0.0285         & 0.4243$\pm$0.0283         & 0.4268$\pm$0.0148         \\
20-20-20-20                            & 0.4178$\pm$0.0176         & 0.4246$\pm$0.0293         & 0.4271$\pm$0.0275         & 0.4287$\pm$0.0289         & 0.4276$\pm$0.0155         \\
50-50-50-50                            & 0.4167$\pm$0.0172         & 0.4267$\pm$0.0259         & 0.4318$\pm$0.0284         & 0.4278$\pm$0.0321         & 0.4297$\pm$0.0174         \\ \Xhline{0.8pt}  
\end{tabular}
\end{table}

To further investigate the impacts of high-dimensional disentangled representations, we meticulously iterate over the dimensions $D_{\textbf{\textit{z}}_{i}},\textbf{\textit{z}}_{i} \in \{\Gamma, \Delta, \Upsilon,\rm E\}$ within the set $\{5,10,15,20,50\}$. Table \ref{SIMU} succinctly captures the varying performance metrics of DRVAE on the dataset Simu($k$) under different dimensions of the factors. The subsequent rows each correspond to the performance of DRVAE when subjected to varying dimensions of disentangled representations. For example, "5-5-5-5" indicates that the dimensions of the four factors are uniformly set to 5. Throughout these experiments, all other parameters are kept constant at their optimal values, with only the factor dimensions being adjusted. We can infer two principal findings from Table \ref{SIMU}: (1) Despite employing a high-dimensional disentangled representation, such as 50, DRVAE markedly surpasses the performance of alternative benchmark methods, with the sole exception of the DPE metric on the Simu3 dataset, where the best performance is attained by GIKS. (2) With an augmentation in the dimensions of the latent factors, DRVAE demonstrates nearly a consistent level of performance across the same dataset. This suggests that the high-dimensional disentangled representations do not overshadow the causal effect of $t$ on $y$.

\begin{figure}[t]
	\centering
	\subfigure[$\alpha^*=0.1$]
	{
		\label{fig:subfig:a}
		\includegraphics[width=0.25\linewidth]{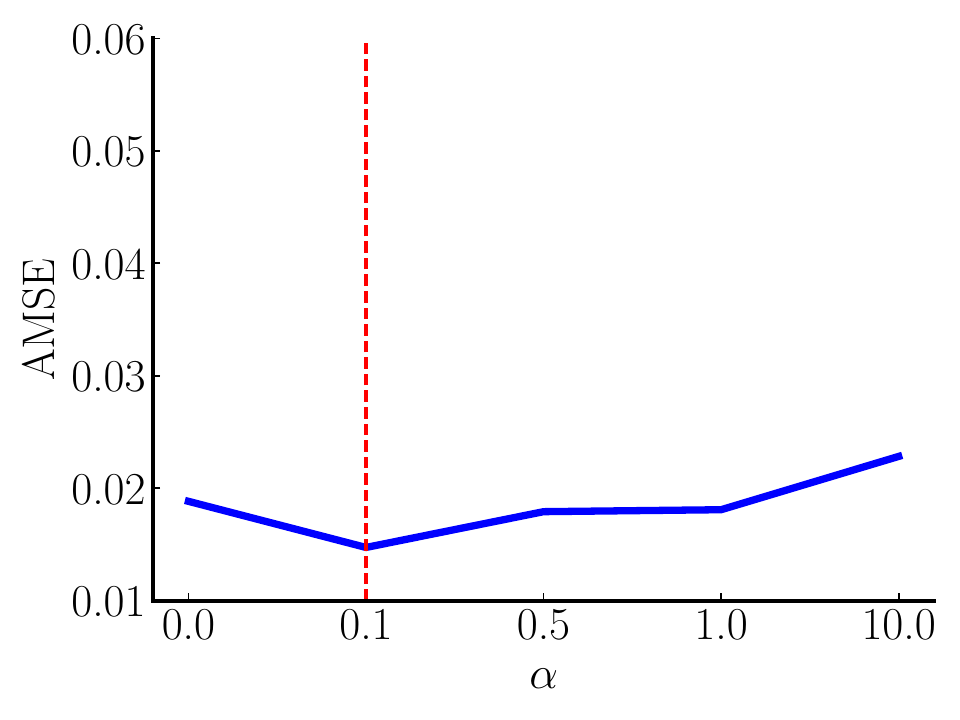}
	}
	\hspace{0.00001\linewidth}
	\subfigure[$\beta^*=1.0$]
	{
		\label{fig:subfig:b}
		\includegraphics[width=0.25\linewidth]{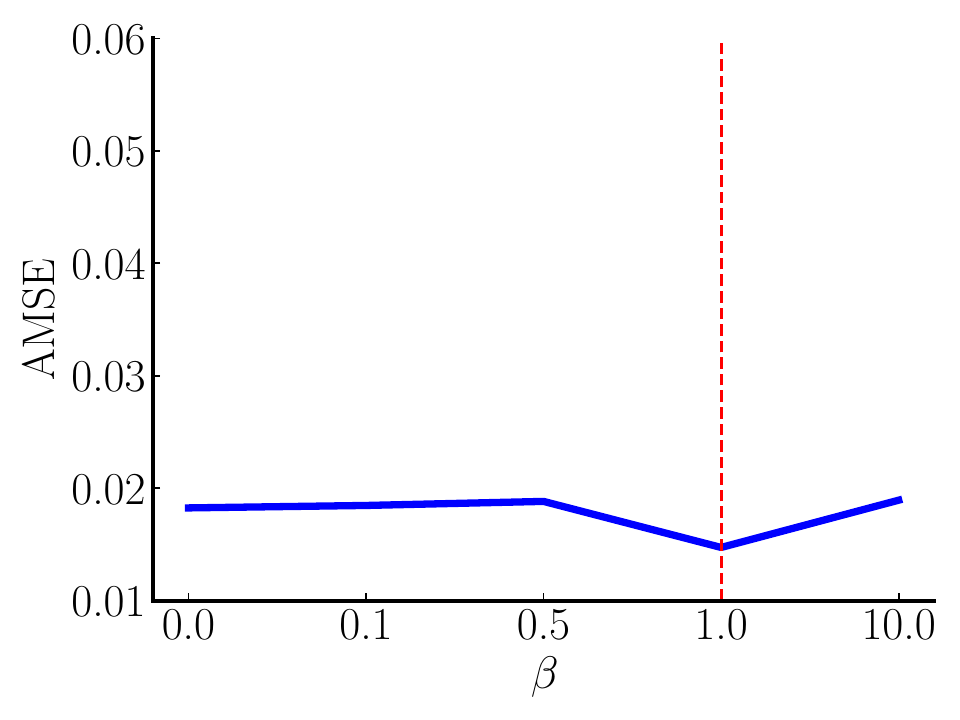}
	}
	\vfill
	\subfigure[$\gamma^*=1.0$]
	{
		\label{fig:subfig:c}
		\includegraphics[width=0.25\linewidth]{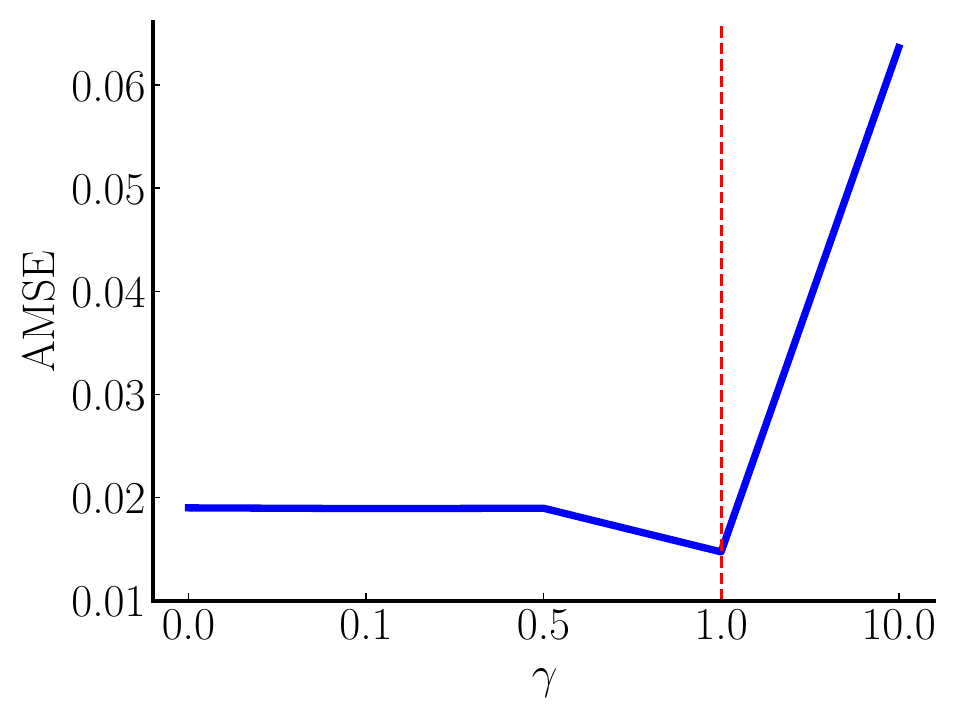}
	}
	\hspace{0.00001\linewidth}
	\subfigure[$\delta^*=0.1$]
	{
		\label{fig:subfig:d}
		\includegraphics[width=0.25\linewidth]{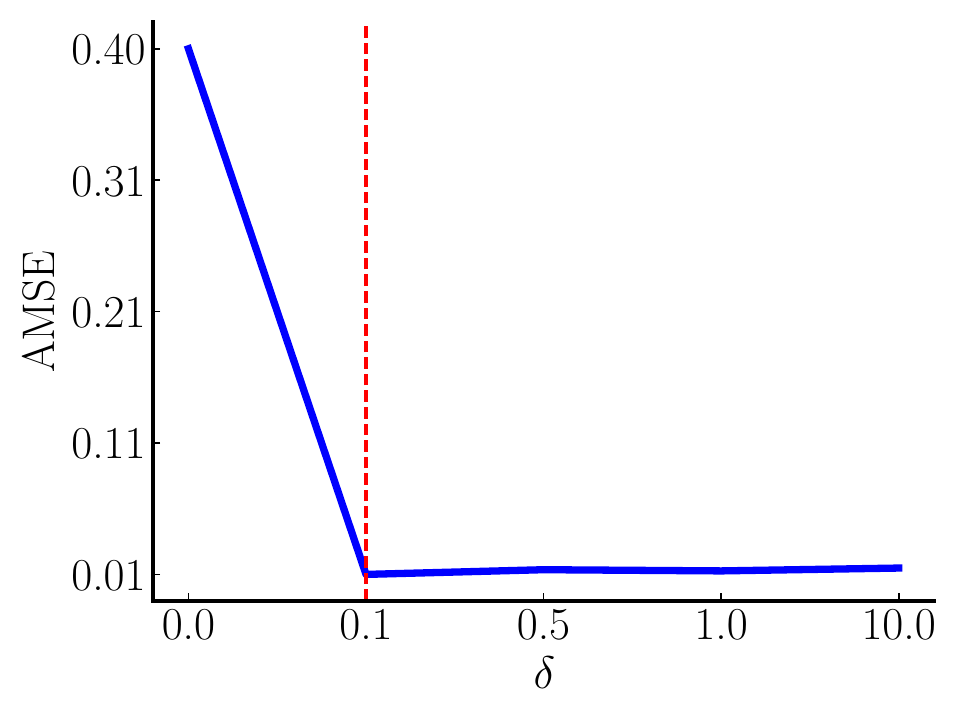}
	}
	\hspace{0.00001\linewidth}
	\subfigure[$\lambda^*=1.0$]
	{
		\label{fig:subfig:d}
		\includegraphics[width=0.25\linewidth]{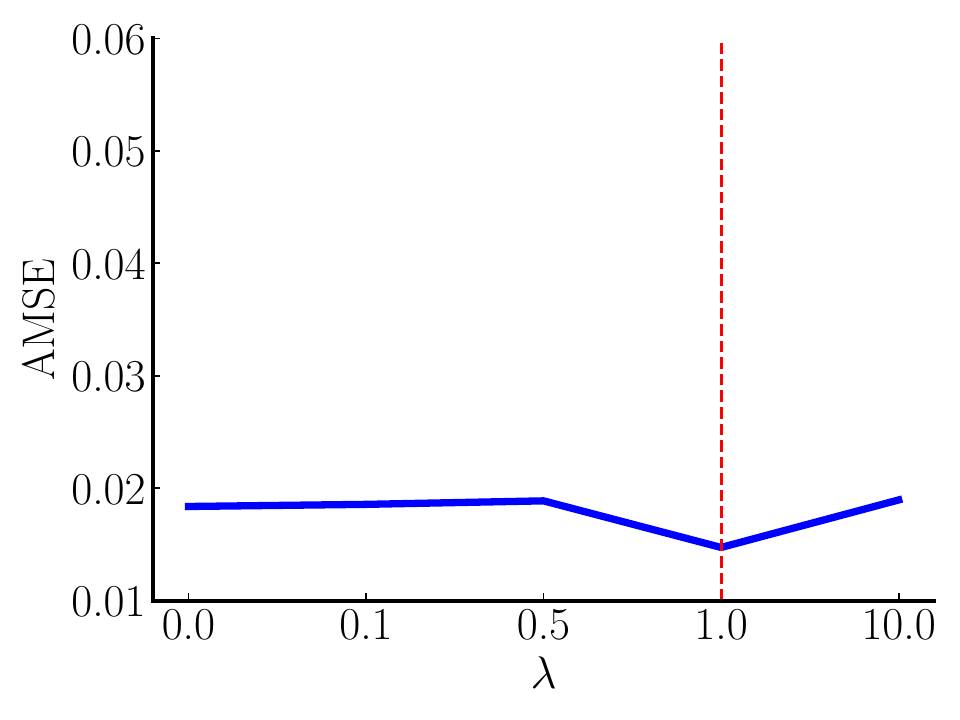}
	}
	\caption{Hyper-parameter sensitivity analysis on $\alpha, \beta, \gamma, \delta$ and $\lambda$. The blue lines show the AMSE of these parameters within the specified range $\{0.0,0.1,0.5,1.0,10.0\}$. The red line indicates the best parameters for the setting.}
	\label{Hyper-parameter}
\end{figure}

\textbf{Impact of the Loss Components on DRVAE Performance.} Considering the complex multi-objective function in DRVAE (Section \ref{Training}), we investigated the impact of each component on the four metrics of the results by varying parameters $\alpha, \beta, \gamma, \delta$ and $\lambda$ within the scope $\{0.0,0.1,0.5,1.0,10.0\}$ in the setting of Simu(1).  Fig. \ref{Hyper-parameter} presents the results of sensitivity analysis for hyperparameters and the optimal values.  It can be observed that $\gamma$ and $\delta$ have a more significant impact on the outcomes, representing the auxiliary losses for predicting $t$ and $y$, respectively. It is worth noting that when $\delta \neq 0$, parameter variations no longer have a significant impact on the results.  DRVAE is not sensitive to parameters $\alpha$, $\beta$ and $\lambda$, but they are necessary.  Loss terms corresponding to $\alpha$ and $\beta$ help train VAE, while the loss term corresponding to $\lambda$ effectively prevents overfitting. With hyper-parameters analysis, we can choose the best hyper-parameters for experiments (see Table \ref{Parameters}).

\begin{table}[h]
	\centering
	\caption{The performance for the DRVAE on the Simu($1$) with different combinations of our loss components. ($\checkmark$ fefers to keeping the component in DRVAE.)}
	\label{para}
	\begin{tabular}{ccccc|llll}
		\toprule 
		$\alpha$ & $\beta$&$\gamma$&$\delta$&$\lambda$& \multicolumn{1}{c}{AMSE} & \multicolumn{1}{c}{$\sqrt{\rm MISE}$} & \multicolumn{1}{c}{$\sqrt{\rm DPE}$} & \multicolumn{1}{c}{$i$-MSE} \\ 
		\midrule 
		&$\checkmark$&$\checkmark$&$\checkmark$&$\checkmark$ & 0.0188$\pm$0.0074       & 0.6174$\pm$0.0088        & \textbf{0.9004$\pm$0.0389}       & 0.4226$\pm$0.0160        \\
	    $\checkmark$&&$\checkmark$&$\checkmark$&$\checkmark$ & 0.0183$\pm$0.0068       & 0.6171$\pm$0.0084        & 0.9078$\pm$0.0393       & 0.4221$\pm$0.0154       \\
		$\checkmark$&$\checkmark$&&$\checkmark$&$\checkmark$ & 0.0190$\pm$0.0077        & 0.6175$\pm$0.0088        & 0.9079$\pm$0.0463       & 0.4228$\pm$0.0156       \\
		$\checkmark$&$\checkmark$&$\checkmark$&&$\checkmark$ & 0.4021$\pm$0.0383       & 0.8301$\pm$0.0129        & 2.1450$\pm$0.0743        & 0.8149$\pm$0.0265       \\
		$\checkmark$&$\checkmark$&$\checkmark$&$\checkmark$&& 0.0184$\pm$0.0070        & 0.6171$\pm$0.0087        & 0.9046$\pm$0.0410        & 0.4222$\pm$0.0157       \\
		\midrule 
	   $\checkmark$&$\checkmark$&$\checkmark$&$\checkmark$&$\checkmark$ & \textbf{0.0147$\pm$0.0075}       & \textbf{0.6134$\pm$0.0091}        & 0.9063$\pm$0.0423       & \textbf{0.4172$\pm$0.0159} \\
		\bottomrule 
	\end{tabular}
\end{table}

Table \ref{para} illustrates the performance of DRVAE under different loss function components. We indicate the absence of a loss function component by setting its corresponding parameter to 0 while keeping other parameters optimal. It can be observed that, except for $\sqrt{\rm DPE}$, DRVAE only demonstrates its best performance when all components of the loss function are present. Among these, the loss term corresponding to $\delta$, i.e. $\mathcal{L} _ { auxiliary} ( y )$, has the greatest impact on the results. This is evident as it directly determines the value of the outcome $y$. The absence of other loss function components has a similar impact on the results.

\section{Conclusion}
\label{Conclusion}
In this paper, we focus on the estimation of continuous treatment effects in observational studies, particularly the estimation of ADRF curves. Previous methods have primarily focused on balancing covariates through the design of network structures or loss functions, thereby overlooking the importance of distinguishing covariates. That is, they treat all covariates as confounding variables. Although many algorithms for disentangling confounding factors have been proposed, they are designed for binary treatment settings and are difficult to extend to continuous cases. Moreover, these methods only decompose covariates into three types of latent factors, with little consideration given to non-causal information such as noise factors mixed in the data. In light of this, we leverage VAE to propose a novel covariate disentanglement representation method for estimating ADRF, coined as DRVAE, which enables us to estimate treatment effects under continuous settings. By representing covariates as instrumental factors, confounding factors, adjustment factors, and external noise factors, we separate and balance the true confounding factors to estimate continuous treatment effects through counterfactual inference. Our experimental section primarily uses the AMSE metric, supplemented by three other types of continuous treatment effect evaluation metrics. Empirical results on the amount of benchmark datasets demonstrate that the performance of the DRVAE surpasses the current state-of-the-art methods.

Compared to existing methods, DRVAE has several advantages. We represent covariates that include a substantial amount of external noise data, which cannot be identified by humans, as four types of factors. By accurately balancing the confounding factors, DRVAE precisely identifies the continuous treatment effects. The disentangled mechanism accurately extracts the true causal information from the raw data and discards non-causal information. Based on the VAE, DRVAE effectively ensures the continuity of the curve while accurately estimating the ADRF curve. Experimental results also show that the dimensions of the four disentangled representations do not obscure the causal effect of treatment on the outcome, making DRVAE more adaptable to any high-dimensional covariates.

Nevertheless, our work has limitations.  DRVAE utilizes the independent representation characteristics of VAE to ensure the independence of the four disentangled representations, and the distributions of them tend towards a standard Gaussian distribution. It means that the absence of a factor does not significantly affect the outcomes. Moreover, we need to theoretically prove that representing covariates as four types of factors will reduce the estimation bias of continuous treatment effects, which will be one of the future research directions.

\section*{Acknowledgment}

This work was supported in part by the National Natural Science Foundation of China under Grants 72071206 and 72231011.

\bibliographystyle{cas-model2-names}
\bibliography{Rreferences}

\section*{Appendix}
\label{Appendix}
\textbf{Synthetic Dataset.} The covariates in the synthetic dataset consist of two parts. Firstly, we generate $\boldsymbol{x}$ as follows: $x_{j}\stackrel{\text{i.i.d.}}{\sim}\text{Unif}[0,1]$, where $x_{j}$ is the $j$-th dimension of $\boldsymbol{x}\in\mathbb{R}^6$. As for the second part, we introduce a portion of continuous noise $\varepsilon_{con}\!\sim\!\mathcal{N}(2, 10)$ and binary noise $\varepsilon_{bin} \!\sim\! Bern(p)$ with $p\!\sim\!\text{Unif}[0,1]$ into the covariates. The dimensions of $\varepsilon_{con}$ are 5$k$, and the dimensions of  $\varepsilon_{bin}$ are 10$k$, where $k=1,2,...,5$ represents the index of the dataset (see Table \ref{3Dataset}). To estimate continuous causal effect, we generate the treatment and outcome following VCNet \cite{VCNet}:
\begin{equation}\begin{aligned}
		\tilde{t}\mid\boldsymbol{x}& \begin{aligned}=\frac{10\sin(\max(x_1,x_2,x_3))+\max(x_3,x_4,x_5)^3}{1+(x_1+x_5)^2}+\sin(0.5x_3)(1+\exp(x_4-0.5x_3))\end{aligned}  \\
		&+x_3^2+2\sin(x_4)+2x_5-6.5+\mathcal{N}(0,0.25), \\
		y\mid\boldsymbol{x},t& \begin{aligned}=\cos(2\pi(t-0.5))\left(t^2+\frac{4\max(x_1,x_6)^3}{1+2x_3^2}\sin(x_4)\right)+\mathcal{N}(0,0.25),\end{aligned} 
\end{aligned}\end{equation}
where $t=(1+\exp(-\tilde{t}))^{-1}$. According to the above mechanism, $x_2, x_5$ are instrumental variables, $x_1, x_3, x_4 $ are confounder variables, $x_6$ is the adjustment variable, and $\varepsilon_{con}$,$\varepsilon_{con}$ are externally noise variables. Consistent with the previous works \cite{VCNet} \cite{TransTEE}, the training set contained 500 samples and the test set contained 200 samples for each repetition.

\textbf{IHDP.} The Infant Health and Development Program (IHDP) \cite{IHDP2} is a randomized controlled trial dataset, initially designed to estimate the causal impact of a binary treatment (home visits of specialists) on infant cognitive test scores. The original dataset contains 747 samples and 25 features. To estimate continuous causal effects, following \cite{VCNet}, we generated the treatment and outcomes using:
\begin{equation}\begin{gathered}
		\widetilde{t}\mid x =\frac{2x_{1}}{(1+x_{2})}+\frac{2\max(x_{3},x_{5},x_{6})}{0.2+\min(x_{3},x_{5},x_{6})}+2\tanh\left(5\frac{\sum_{i\in S_{dis,2}}(x_{i}-c_{2})}{|S_{dis,2}|}\right)-4+\mathcal{N}(0,0.25), \\
		y\mid\boldsymbol{x},t =\frac{\sin(3\pi t)}{1.2-t}\left(\mathrm{tanh}\left(5\frac{\sum_{i\in S_{\mathrm{dis,~}1}}(x_{i}-c_{1})}{|S_{\mathrm{dis,~}1}|}\right)+\frac{\exp(0.2(x_{1}-x_{6}))}{0.5+5\min(x_{2},x_{3},x_{5})}\right)+\mathcal{N}(0,0.25), 
\end{gathered}\end{equation}
where $t=(1+\exp(-\tilde{t}))^{-1},S_{\mathrm{con}}=\{1,2,3,5,6\}$ is the index set of continuous features, $S_{\mathrm{dis},1} = \{4,7,8,9,10,11,12,13,\\
14,15\}$, $S_{\mathrm{dis},2}=\{16,17,18,19,20,21,22,23,24,25\}$, $S_{\mathrm{dis},1}\cup S_{\mathrm{dis},2}=[25]-S_{\mathrm{con}}$. Here $c_{1}=\mathbb{E}\frac{\sum_{i\in S_{\mathrm{dis},1}}x_{i}}{|S_{\mathrm{dis},1}|},c_{2}=\mathbb{E}\frac{\sum_{i\in S_{\mathrm{dis},2}}x_{i}}{|S_{\mathrm{dis},2}|}.$ According to the mechanism, $S_{\mathrm{dis},2}$ are instrumental variables, $S_{\mathrm{con}}$ are confounder variables, and $S_{\mathrm{dis},1}$ are the adjustment variables.

\textbf{News.} The original News dataset comprises 3000 item samples. We adopt a data preprocessing procedure akin to that of \cite{VCNet}. Given that the dataset is employed to estimate the effect of binary treatment, we adhere to the approach outlined in \cite{SCIGAN} \cite{VCNet} to generate continuous treatments and their corresponding outcomes as follows:
\begin{equation}
	\label{Enews}
	\begin{aligned}
		&\boldsymbol{v}_{i}'\sim \mathcal{N}(\mathbf{0},\mathbf{1}),\\ &\boldsymbol{v}_{i}=\boldsymbol{v}_{i}^{\prime}/\left\|\boldsymbol{v}_{i}^{\prime}\right\|_{2}\mathrm{~for~}i=\{1,2,3\},\\
		&t \sim \mathrm{Beta}\left(2,\left|\frac{\boldsymbol{v}_3^\top\boldsymbol{x}}{2\boldsymbol{v}_2^\top\boldsymbol{x}}\right|\right),\\
		&y^{\prime}\mid\boldsymbol{x},t=\exp\left(\frac{\boldsymbol{v}_2^\top\boldsymbol{x}}{\boldsymbol{v}_3^\top\boldsymbol{x}}-0.3\right),\\
		&y\mid\boldsymbol{x},t=2\left(\max(-2,\min(2,y^{\prime}))+20\boldsymbol{v}_1^\top\boldsymbol{x}\right)*\left(4\left(t-0.5\right)^2*\sin\left(\frac{\pi}{2}t\right)\right)+\mathcal{N}(0,0.5).
\end{aligned}
\end{equation}

In contrast to other two datasets, the generative process of the News incorporates a higher level of stochasticity. Initially, the incorporation of $\boldsymbol{v}_{i}$ replaces the deterministic function with a probabilistic distribution $Beta$ for the generation of $t$. Furthermore, the generation of $y$ is augmented with noise $\mathcal{N}(0,0.5)$. Consequently, while our approach may exhibit suboptimal performance on certain metrics, it maintains a superior performance on the principal metric, AMSE, outperforming other methods (see Table \ref{News}).

\end{document}